\pdfoutput=1

\documentclass[11pt]{article}
\usepackage{amsmath} 
\usepackage{booktabs} 
\usepackage{multirow}
\usepackage[final]{acl}

\usepackage{times}
\usepackage{latexsym}

\usepackage[T1]{fontenc}

\usepackage[utf8]{inputenc}

\usepackage{microtype}

\usepackage{inconsolata}

\usepackage{graphicx}
\usepackage{subcaption}
\usepackage{caption}
\usepackage{dblfloatfix}
\usepackage{url}
\usepackage{paralist,amsmath,amssymb,adjustbox,booktabs}
\usepackage{authblk}

\title{Images Speak Louder than Words: Understanding and Mitigating Bias in Vision-Language Model from a Causal Mediation Perspective}

\author[1]{Zhaotian Weng}
\author[1]{Zijun Gao}
\author[2]{Jerone Andrews}
\author[1]{Jieyu Zhao}

\affil[1]{University of Southern California}
\affil[2]{Sony AI}

\affil[ ]{\texttt{\{wengzhao, jieyuz\}@usc.edu}, \texttt{zijungao@marshall.usc.edu}, \texttt{Jerone.Andrews@sony.com}}

\begin{document}
\maketitle
\begin{abstract}

Vision-language models (VLMs) pre-trained on extensive datasets can inadvertently learn biases by correlating gender information with specific objects or scenarios. Current methods, which focus on modifying inputs and monitoring changes in the model’s output probability scores, often struggle to comprehensively understand bias from the perspective of model components. We propose a framework that incorporates causal mediation analysis to measure and map the pathways of bias generation and propagation within VLMs. Our framework is applicable to a wide range of vision-language and multimodal tasks. In this work, we apply it to the object detection task and implement it on the GLIP model. This approach allows us to identify the direct effects of interventions on model bias and the indirect effects of interventions on bias mediated through different model components. Our results show that image features are the primary contributors to bias, with significantly higher impacts than text features, specifically accounting for approximately 33\% and 13\% of the bias in the MSCOCO and PASCAL-SENTENCE datasets, respectively. Notably, the image encoder's contribution surpasses that of the text encoder and the deep fusion encoder. Further experimentation confirms that contributions from both language and vision modalities are aligned and non-conflicting. Consequently, focusing on blurring gender representations within the image encoder which contributes most to the model bias, reduces bias efficiently by approximately 22.03\% and 9.04\% in the MSCOCO and PASCAL-SENTENCE datasets, respectively, with minimal performance loss or increased computational demands.

\end{abstract}

\section{Introduction}
Vision-language models have shown promising results in tasks such as classification~\cite{li2023blip,jia2021scaling,radford2021learning}, image search~\cite{radford2021learning,li2023blip}, and object detection~\cite{kuo2023openvocabulary,li2022grounded} by training on large-scale image-text pairs to understand the correspondences between cross-modal image features and language features. Models trained on extensive datasets exhibit excellent zero-shot capabilities~\cite{radford2021learning,yu2022coca,li2022grounded,zhang2022glipv2} but also risk discovering and exploiting societal biases present in the underlying image-text pair corpora, potentially introducing bias that leads to social unfairness ~\cite{zhao2017men}. The revelation, measurement, and understanding of biases within models ~\cite{zhou-etal-2022-vlstereoset,zhang2022counterfactually,lee2023survey,vig2020investigating} has sparked widespread research interest and are crucial for bias mitigation ~\cite{zhang2022counterfactually,seth2023dear,dehdashtian2024fairerclip}. However, most contemporary methods, derived from language models, lack standardized metrics for evaluating bias and primarily assess the correlation between the outputs of classifiers and external attributes ~\cite{zhang2022counterfactually}. ~\citet{barrett2019adversarial} noted that interpretations based on classifier outputs can be factually inaccurate and not generalizable. While these methods can highlight the impacts of certain contributions on model outputs, they (1) fail to comprehend the generation and flow of bias within the model and (2) do not understand the causal roles of model components in the generation and propagation of bias. Consequently, they are not able to provide clear guidance on how to effectively mitigate bias at the model level.

In this work, we propose a standardized framework to measure bias in VLMs, providing a comprehensive understanding of how bias flows within the entire model structure. Specifically, we use the GLIP model~\cite{li2022grounded} as a case study, focusing on gender bias in the task of object detection, which is a predominant and challenging problem in computer vision.
We conduct the analysis on both the MS-COCO~\cite{lin2014microsoft} and PASCAL-SENTENCE~\cite{rashtchian2010collecting} datasets.
We observe that GLIP model exhibits unbalanced inference capabilities on different genders, with certain indoor object categories like pets more likely to be associated with females and outdoor objects like vehicles with males. 
To holistically understand how the bias flows in the model, we adapt causal mediation analysis ~\cite{vig2020investigating} to VLMs, providing a finer-grained study of contributions from different model components.
We find that, among the different model components (text module, image module, and fusion module that combines them), the image module contributes the most to the model's bias -- over twice as much as the text module. In the MSCOCO and PASCAL-SENTENCE datasets, image features accounted for approximately 32.57\% and 12.98\% of the bias generated, respectively, compared to approximately 15.48\% and 5.64\% by text features. Also, the interaction and updating process between image and text features during the deep fusion process significantly impacts bias production, accounting for about 56\% of the contributions in the image and text encoders. Furthermore, by integrating interventions across different modules, we discovered that their contributions to bias are aligned and do not conflict, allowing us to prioritize bias mitigation efforts within the image encoder, which is the most substantial contributor to bias.
Based on the results, we propose to effectively mitigate the bias in VLMs: reducing biases from the image module can successfully reduce bias by approximately 22.03\% on the MSCOCO dataset and 9.04\% on the PASCAL-SENTENCE dataset, compared to a reduction of approximately 7.81\% and 1.18\% in the text module.
In summary, the contributions of our work are: 
\begin{compactitem}
    \item We provide a comprehensive evaluation of the bias in VLMs, with an understanding of the  contribution from each model module, which is missing in the literature.
    \item We analyze the correlation between the biases from different modules and discover that the bias in different modules are aligned and do  not conflict with each other.

    \item We propose an effective bias mitigation strategy to reduce the bias from the module that contributes most to the model bias when facing a limited budget. 
    
\end{compactitem}

\section{Related Work}

In recent years, vision-language models (VLMs) have experienced rapid advancements. The latest developments in VLMs often employ a dual-stream architecture that separately encodes text and images~\cite{kim2021vilt}, and these are then merged and aligned to facilitate cross-modal understanding of visual and linguistic features ~\cite{radford2021learning}. Furthermore, some studies treat the joint training of images and text as a phrase localization process, aiming to better align and integrate visual and linguistic features ~\cite{li2022grounded}. Typically, these models are trained on image-text pairs from datasets such as MSCOCO ~\cite{lin2014microsoft}, VQA ~\cite{antol2015vqa}, OpenImages ~\cite{kuznetsova2020open}, and Flickr30k Entities ~\cite{7410660}, achieving impressive results in various downstream tasks including image classification ~\cite{radford2021learning}, image generation, visual question answering ~\cite{li2018vqa, antol2015vqa}, and image captioning ~\cite{lu2019vilbert, alayrac2022flamingo}.

Alongside their development, the societal biases exhibited by VLMs have also attracted significant attention. These models often reflect societal stereotypes and may even amplify such biases ~\cite{zhou-etal-2022-vlstereoset}. Most contemporary research addressing bias in VLMs has borrowed methodologies from language model studies. For instance, ~\citet{srinivasan-bisk-2022-worst} utilized a language masking model to explore gender biases by using templates containing a specific entity and analyzing the probabilities of masked entities ~\cite{kurita2019measuring}. Some researchers have examined biases through the comparison of factual and counterfactual inputs, with ~\citet{zhang2022counterfactually} investigating biases by examining predicted probabilities from both factual and counterfactual inputs, and ~\citet{howard2024uncovering} using the Perspective API to score predictions derived from such inputs to study model biases.

However, existing evaluation methods primarily observe changes in the probability scores of model outputs following interventions on input samples. This approach limits our understanding of the underlying causes of bias generation and propagation within model components ~\cite{barrett2019adversarial}. Therefore, we propose a standardized framework for evaluating bias in vision-language tasks and introduce causal mediation analysis ~\cite{robins1992identifiability, pearl2022direct, vig2020investigating} within the context of vision-language models. This methodology helps us understand the pathways of bias generation and propagation from the input level to model components.

\section{Bias Measurement and Understanding in VLM}

In this section, we propose a bias evaluation metric to assess the bias of VLM in the object detection task. By applying causal mediation analysis, we quantify the contribution on bias from various components within the model pipeline which helps us trace the origins and propagation of bias throughout the model pipeline. Additionally, we investigate the interactions between different modalities to understand how they collectively influence model bias which will be used as guidance for bias mitigation later.

 \subsection{Bias Evaluation Metrics}

 In the literature, there have been  various methodologies proposed to measure bias, including notable contributions from ~\citet{zhao2017men},~\citet{wang2021directional} and ~\citet{zhao2023men}. These studies often assess bias amplification by comparing statistics between the training dataset and model outputs, where the models are trained and tested on similarly distributed datasets. In contemporary settings, most VLMs undergo training on extensive collections of image and text corpora. In real-world applications, users may fine-tune a model on a dataset specific to a downstream task. The combination of fine-tuning data and pre-training data can introduce noise, complicating the statistics of previously mentioned bias evaluations. Additionally, many pre-training datasets used for large-scale models are either difficult to access or require significant computational resources for analysis, making existing evaluations challenging to deploy in modern settings.
 
 Notably, recent advancements in VLM have demonstrated impressive zero-shot performance, enabling models to make accurate predictions on benchmark datasets without any fine-tuning ~\cite{radford2021learning,yu2022coca,li2022grounded,zhang2022glipv2}.
 In our study, we explore a zero-shot scenario where VLMs are directly tasked with making predictions on a benchmark dataset without any fine-tuning.

Drawing inspiration from observations in~\citet{zhao2017men}, where females typically correlate more closely with indoor objects than males, we introduce the definition of \textsc{Bias\textsubscript{VL}} which captures model's underlying correlations between sensitive attributes (e.g., genders) and objects:

\begin{align}
    \text{\textsc{Bias\textsubscript{VL}}} := & \sum_{\text{object}} \left| \mathcal{C}(\text{object, male}) \right. \notag \\
    & \left. - \mathcal{C}(\text{object, female}) \right|
    \label{eq:indirect_bias}
\end{align}
where ${\mathcal{C}(x,y)}$ measures the correlation between $x$ and $y$. In our case, we use a false positive rate (FPR) to describe the correlation, which measures how often one specific gender $y$ can trigger a model to incorrectly predict one object $x$ in the image.~\footnote{Following existing work, we also consider binary gender in this study.}

\subsection{Causal Mediation Analysis Method}

\begin{figure}[t]
  \centering
  \includegraphics[width=\columnwidth]{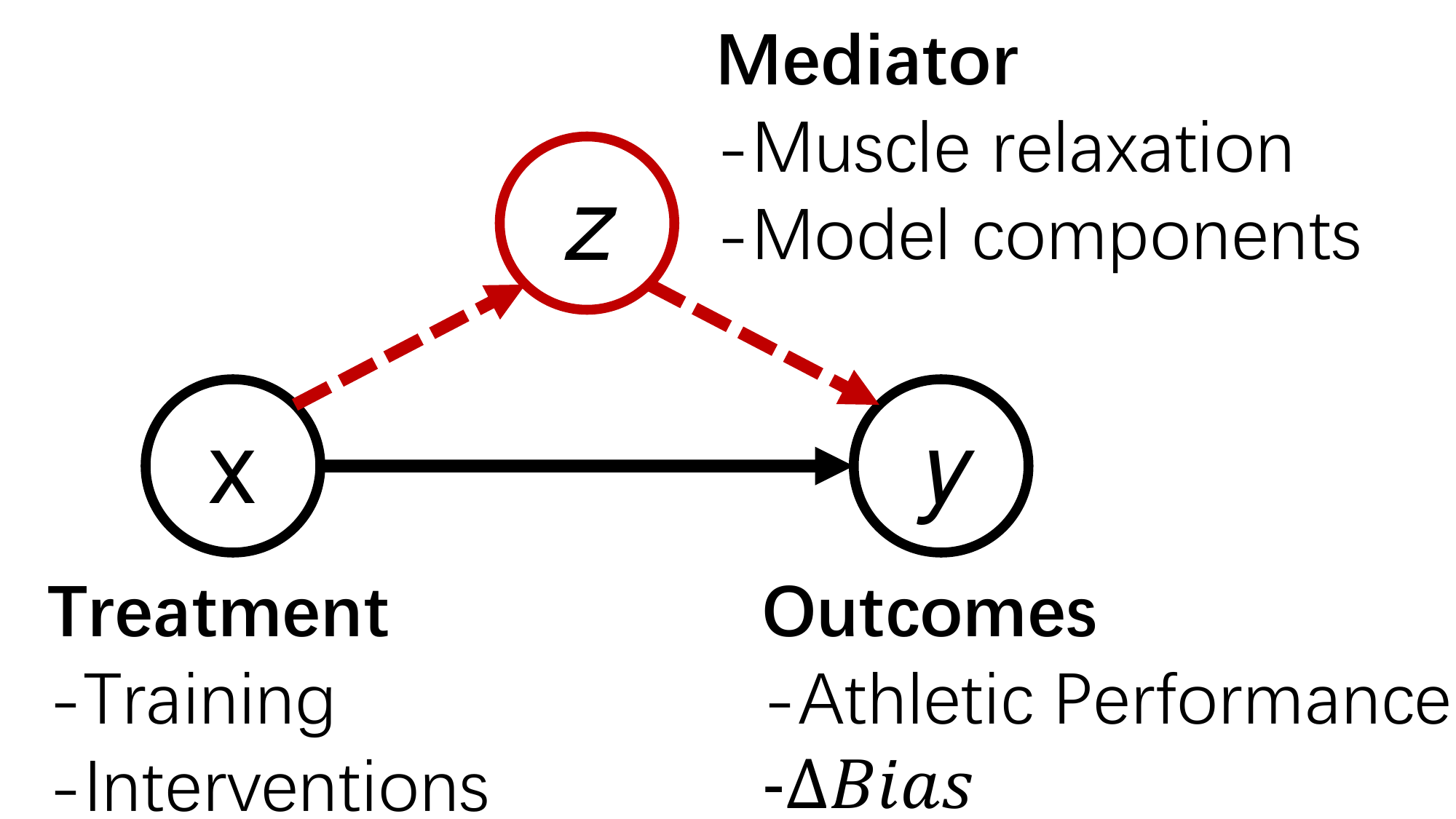}
  \caption{Causal Mediation Analysis example. In this example, strength training serves as the treatment, aiming to improve athletic performance, while muscle relaxation acts as the mediator that indirectly affects the athletic performance. In our study, interventions in the input module directly affect gender bias in model predictions, while model components such as specific layers or neurons can mediate this effect.}
  \label{fig:mediator}
\end{figure}

Causal mediation analysis measures how a treatment effect influences an outcome either directly or indirectly through a mediator variable ~\cite{robins1992identifiability,vig2020investigating,robins2003semantics,pearl2022direct}. An illustrative example is shown in Figure~\ref{fig:mediator}, where athletes engage in strength training to improve athletic performance. After training, they need muscle relaxation to alleviate soreness, which also impacts performance. Thus, strength training can have a direct effect on athletic performance through its intended mechanisms and an indirect effect through muscle relaxation.

In our study, the treatment consists of interventions on the input module, while the mediator could be any model component or finer-grained layer or neuron we are interested in and the outcome is the change in gender bias in the model's prediction results. Therefore, we define three types of intervention:
a) \texttt{replace-gender}, which replaces the gender word \textit{man} or \textit{woman} to a gender-neutral word \textit{person} in the text of the input module; b) \texttt{mask-gender}, where pixels corresponding to a person in the image module are masked, thus removing gender information from the input images; and c) \texttt{null}, which leaves the original text and image modules unchanged.
\begin{figure*}[!] 
    \centering
    \begin{subfigure}[t]{0.22\textwidth}
        \centering
        \includegraphics[width=\textwidth]{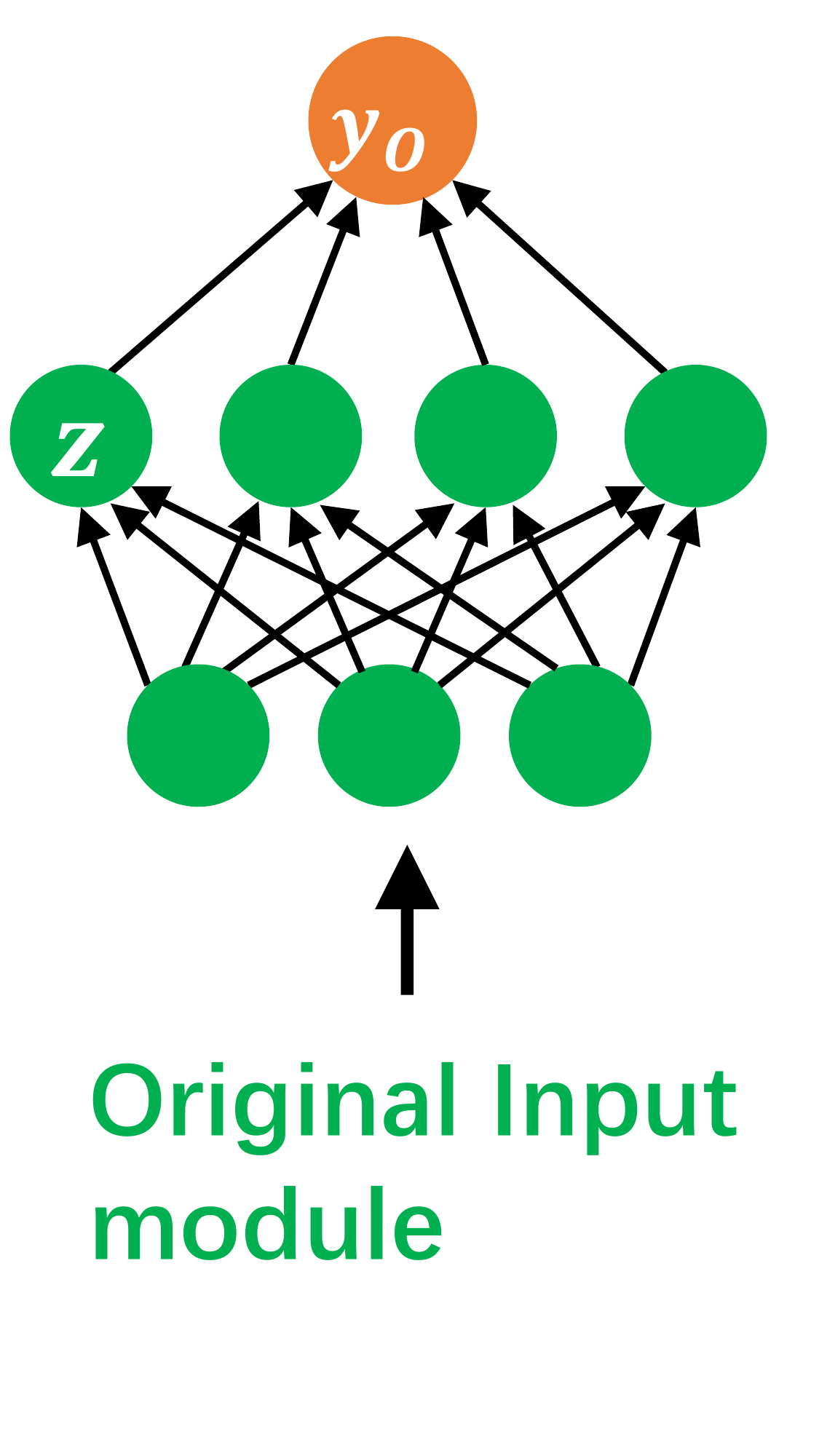}
        \caption{Baseline}
        \label{fig:baseline_effect}
    \end{subfigure}
    \hfill
    \begin{subfigure}[t]{0.22\textwidth}
        \centering
        \includegraphics[width=\textwidth]{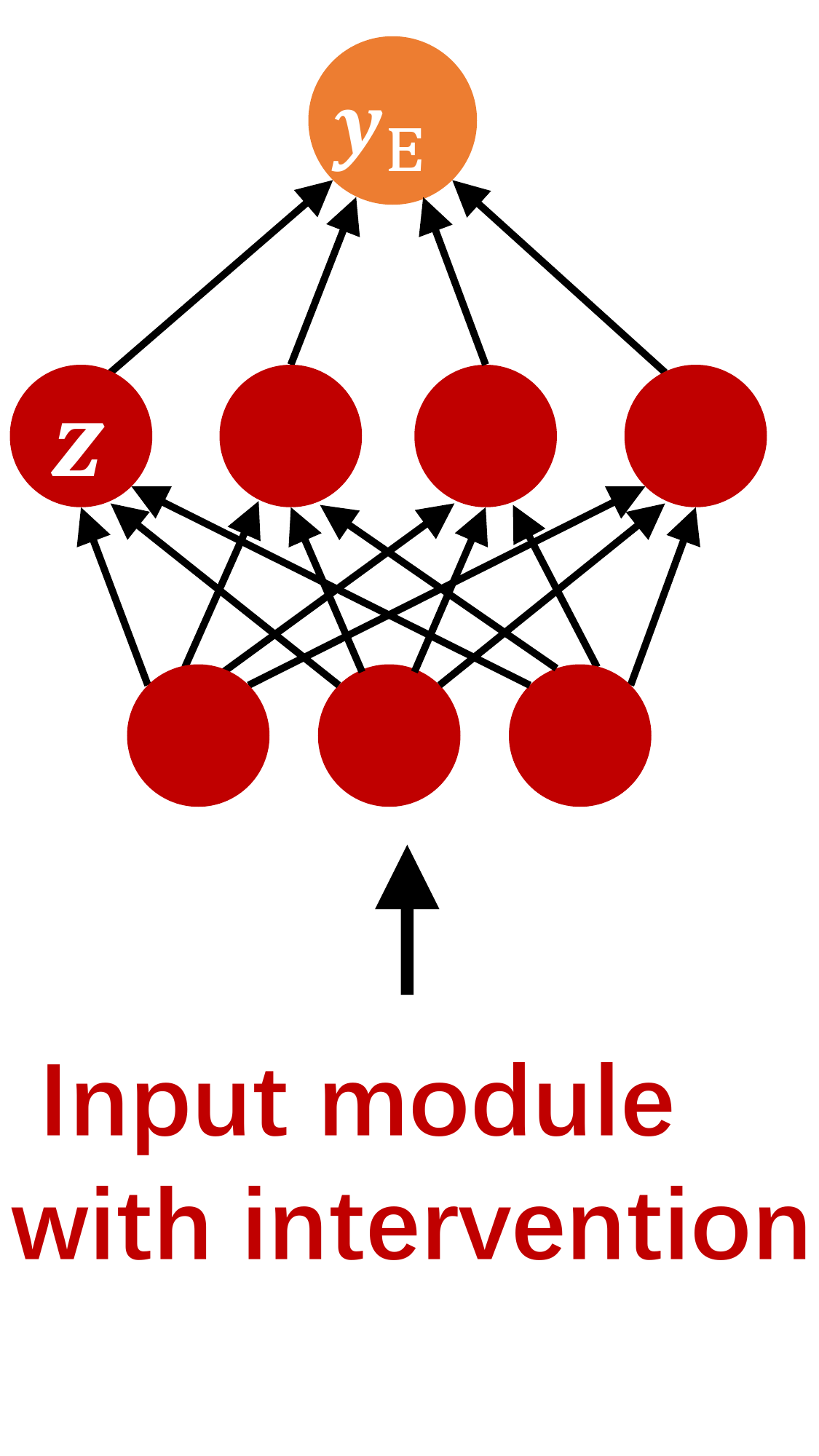}
        \caption{Intervention Effect\\ $E := y_E - y_O$}
        \label{fig:total_effect}
    \end{subfigure}
    \hfill
    \begin{subfigure}[t]{0.22\textwidth}
        \centering
        \includegraphics[width=\textwidth]{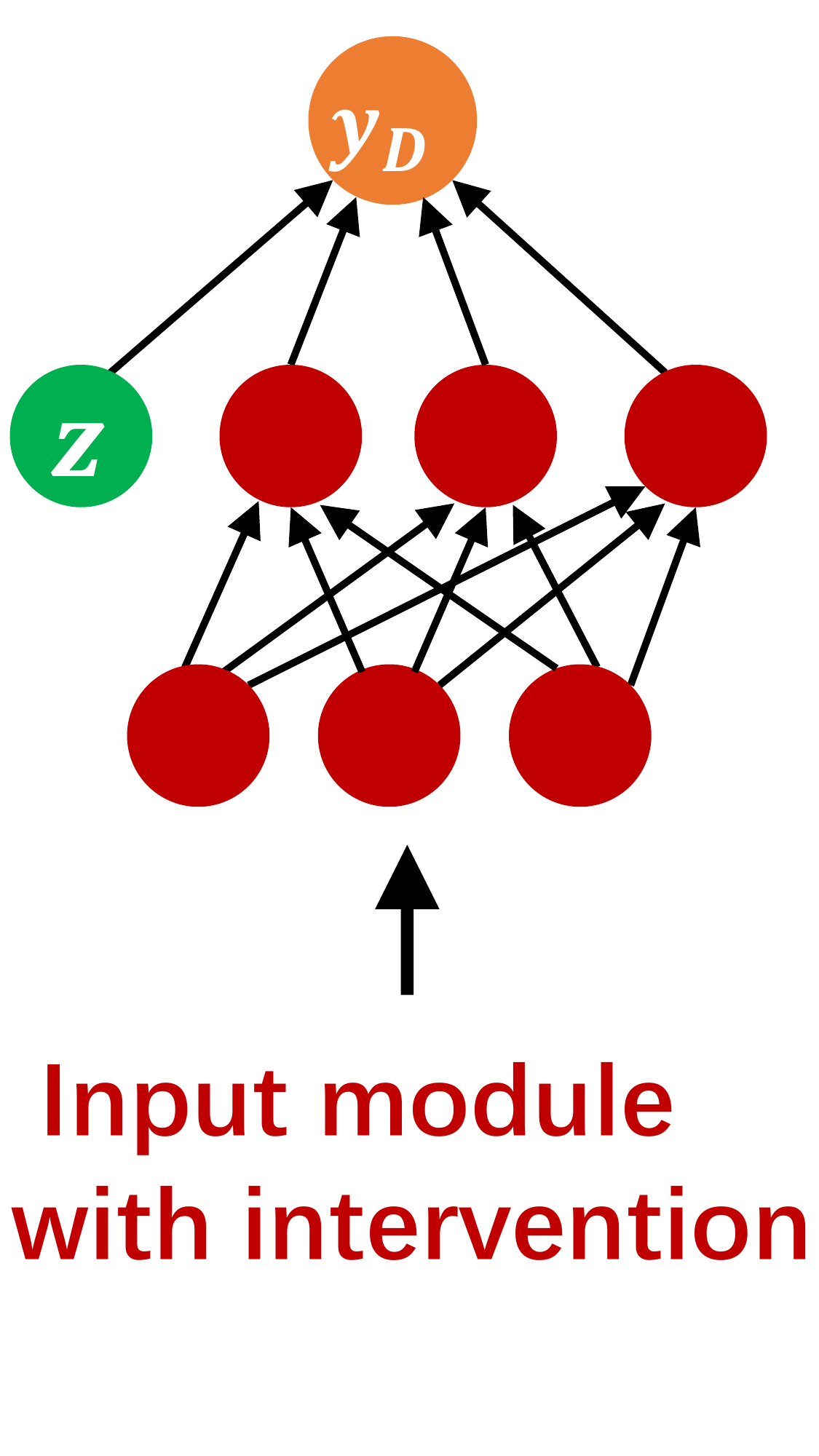}
        \caption{Direct Effect\\$DE := y_D - y_O$}
        \label{fig:direct_effect}
    \end{subfigure}
    \hfill
    \begin{subfigure}[t]{0.22\textwidth}
        \centering
        \includegraphics[width=\textwidth]{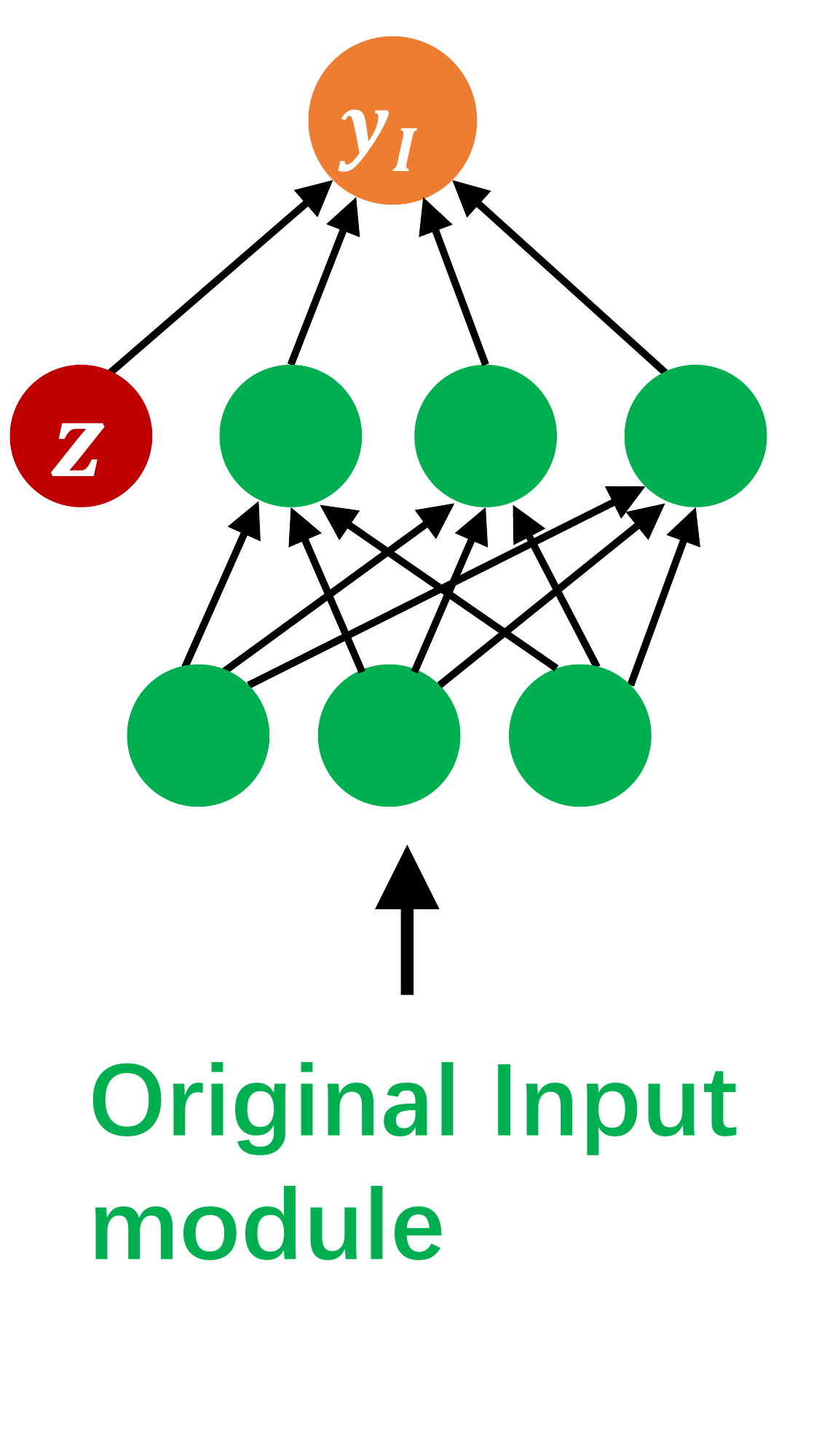}
        \caption{Indirect Effect\\$IE := y_I - y_O$}
        \label{fig:indirect_effect}
    \end{subfigure}
    \caption{Bias understanding with causal mediation analysis. In the diagram, \textit{z} represents the mediator, and $y_O$, $y_E$, $y_D$, $y_I$ represent the bias values of the model's output under various interventions. The \textbf{\textit{intervention effect}} quantifies the change in the bias score under the specified intervention; the \textbf{\textit{direct effect}} quantifies the change in bias score resulting from an intervention in the input module while maintaining the mediator in the state of a \texttt{null} intervention; the \textbf{\textit{indirect effect}} measures the change in the bias score when the input module remains unchanged, but the mediator is set to the state of a specific intervention. 
    }
    \label{fig:causal_mediation_example}
\end{figure*}

We perform causal mediation analysis on the GLIP model by introducing interventions in the input module and observing changes in \textsc{Bias\textsubscript{VL}} values defined in Eq.\eqref{eq:indirect_bias}.
 Following ~\citet{vig2020investigating}, we define the \textbf{\textit{Direct Effect (DE)}} as changes in the \textsc{Bias\textsubscript{VL}} score when the intervention is applied to the input module while the mediator (model components) remains in the `\texttt{null}' state of intervention. The \textbf{\textit{Indirect Effect (IE)}} represents changes in the bias score when the input module is fixed, but the mediator is set in the state of a certain intervention.
We can select any model structure of interest as the mediator and choose \texttt{`mask-gender', `replace-gender'}, or combinations of them as interventions in the input module (Figure~\ref{fig:causal_mediation_example}).

\section{Experimental Setup of Bias Measurement and Understanding}
\begin{figure*}[t]
  \centering
  \includegraphics[width=0.8\textwidth]{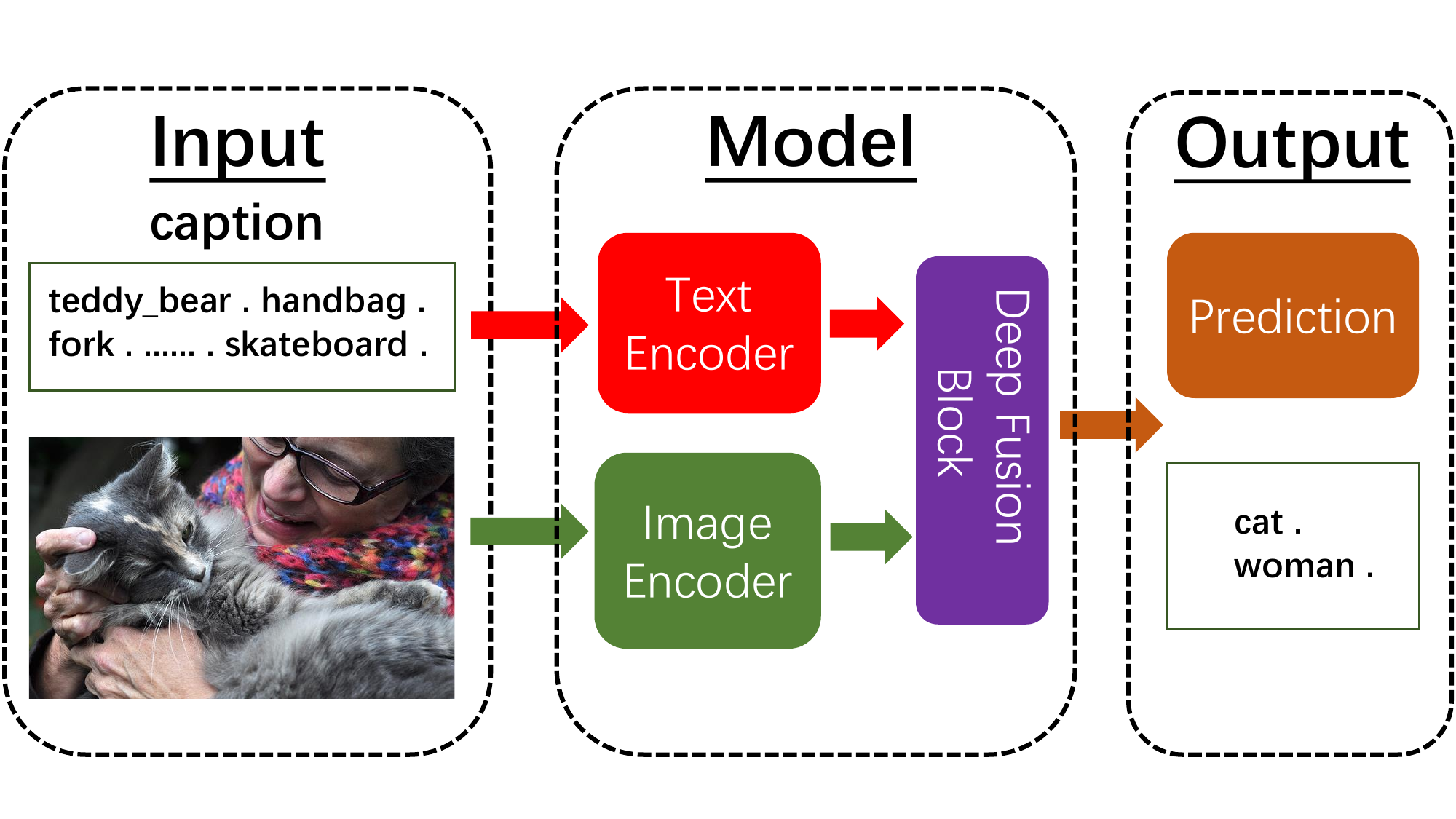}
  \caption{GLIP object detection pipeline. The input to the GLIP model consists of an image and a caption, which includes a list of possible categories separated by commas.}
  \label{fig:pipeline}
\end{figure*}

\paragraph{Model} For the object detection task, we employed the GLIP model pre-trained on the O365, GoldG, CC3M, and SBU datasets~\cite{li2022grounded}. The model consists of an image module, a text module, and a deep-fusion module that updates and aligns image features and text features. For object detection, the GLIP model makes predictions based on the given image and a text input, which is a list of possible categories separated by commas. This input format, as highlighted by ~\citet{li2022grounded}, has demonstrated good performance. For example, in the case of the COCO dataset, the text input along with an image will be a text prompt like "teddy\_bear . handbag . fork ...... . baseball\_glove . skateboard .", as illustrated in Figure~\ref{fig:pipeline}.
\paragraph{Dataset}Our experiments were conducted on the MSCOCO and PASCAL-SENTENCE datasets. For MSCOCO, we follow the  setting in ~\citet{zhao2017men}, where we only consider 66 objects that appear with man or woman more than 100 times in the training data. For the PASCAL-SENTENCE dataset, which includes 20 categories but lacks gender labels, we annotated gender based on the five captions associated with each image. An image is labeled as male if any caption mentions ``male, males, man, men, boy, boys'' and as female if any caption mentions ``female, females, woman, women, girl, girls''. Images that do not include any person or mention both genders were excluded.

\paragraph{Interventions on image encoder and text encoder}

Initially, we implement  \texttt{replace-gender} and \texttt{mask-gender} interventions on the inputs respectively without any alterations to the model components. By monitoring the changes in the values of \textsc{Bias\textsubscript{VL}},
the individual impacts of image and text inputs on gender bias within the input module were assessed. Subsequently, we conducted a detailed causal mediation analysis on the text encoder and image encoder. Previous studies ~\cite{vaswani2017attention,vig2019analyzing} have emphasized the critical roles that attention heads and layers play in deep learning models. Therefore, we selected attention heads from these two modules as mediators. Specifically, we chose the attention heads within a specific layer, along with those in all preceding layers as mediators, and conducted experiments from shallow to deep layers. This analysis aimed to identify whether the text encoder or image encoder contributes more significantly to gender bias and to determine which layers in the model are principally responsible for bias generation. It also sought to understand how bias flows and accumulates across different layers within the encoders. Then, we selected a combination of attention layers from both the image encoder and text encoder as mediators to observe changes in bias and compare these results with previous findings, exploring whether different modalities reinforce bias or conflict in the direction of bias.

\paragraph{Interventions on deep-fusion encoder} In the deep fusion encoder, where image and text features dynamically interact and are updated, we implement \texttt{replace-gender} and \texttt{mask-gender} interventions in the input module to control the state of image and text features within the deep fusion module. We also select the attention heads within a specific layer and all preceding layers' attention heads as the mediator for conducting causal mediation analysis. By observing changes in the values of \textsc{Bias\textsubscript{VL}}, we explore how image and text features individually affect the deep fusion process and subsequently influence bias generation.

\section{Results}
\subsection{Bias Measurement}

We present the results of \textsc{Bias\textsubscript{VL}} in Table~\ref{tab:indirect_bias}, for the MSCOCO dataset, without any intervention on the inputs, the \textsc{Bias\textsubscript{VL}} measured was 1.434. To highlight the significance of this bias, we randomly divided subsets composed of male images into two equal parts, achieving an \textsc{Bias\textsubscript{VL}} of 0.278. Similarly, dividing female image subsets randomly resulted in an \textsc{Bias\textsubscript{VL}} of 0.359. Both results are significantly lower than 1.434, and comparable results were observed with the PASCAL-SENTENCE dataset, as detailed in Table~\ref{tab:indirect_bias}.  The results in the random division  demonstrate that  a model with balanced inference capabilities across a dataset would yield minimal \textsc{Bias\textsubscript{VL}} values when divided into equal subsets (i.e., the gender stays the same). However, when model predictions are influenced by attributes such as gender, splitting the dataset based on such attributes leads to higher \textsc{Bias\textsubscript{VL}} values.

\begin{figure}[ht]
  \centering
  \includegraphics[width=\columnwidth]{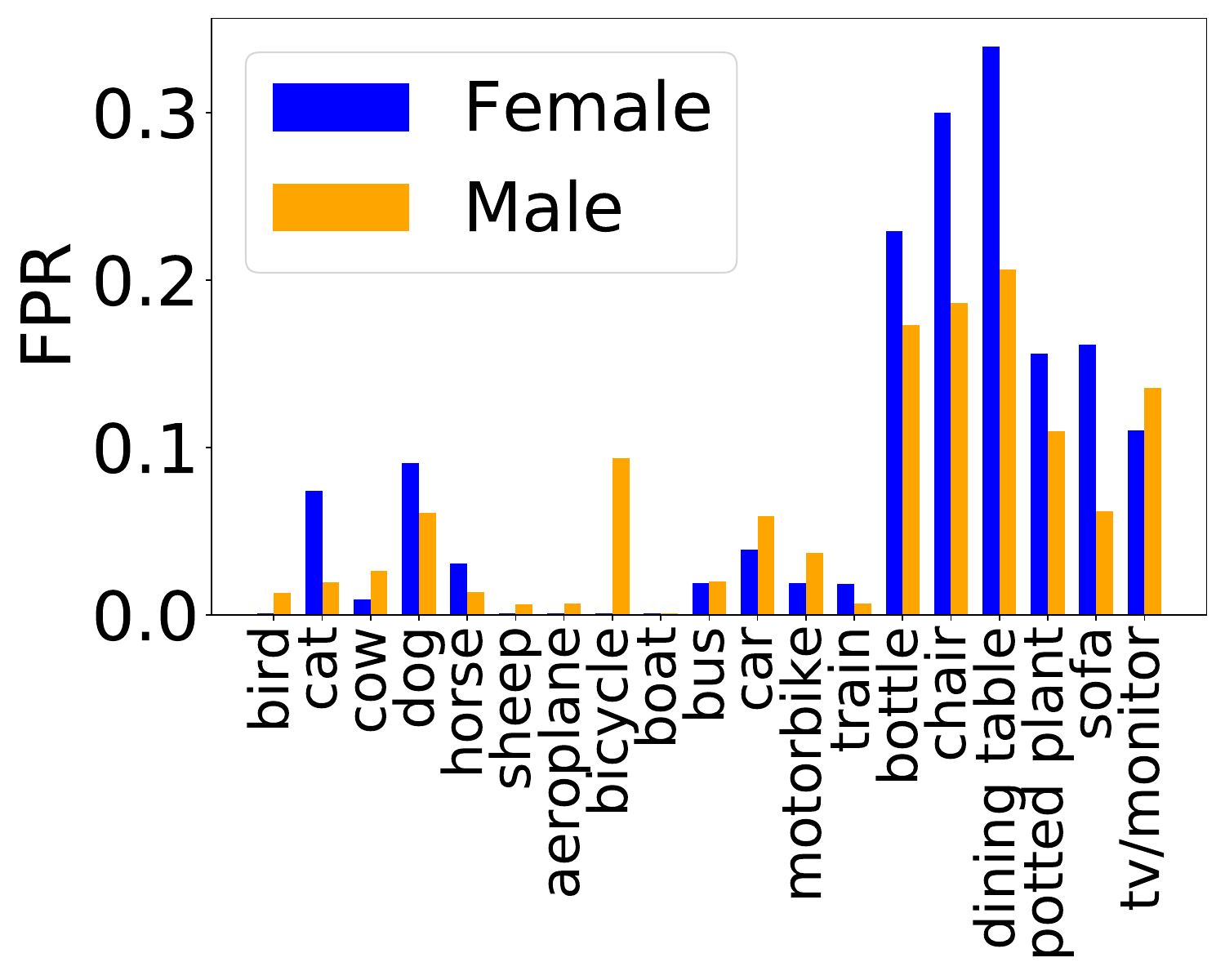}
  \caption{False Positive Rate (FPR) for various objects in the PASCAL-SENTENCE dataset. For most pets and indoor objects, the FPR is higher in images of females than in those of males; conversely, for most outdoor objects such as vehicles, the FPR is higher in images of males than in those of females. These results indicate that females correlate more closely with indoor objects than males.}
  \label{fig:pascal_fpr}
\end{figure}

We also provide detailed statistics of False Positive Rate (FPR) scores for various objects in the PASCAL-SENTENCE dataset, presented in Figure~\ref{fig:pascal_fpr}. Our statistics reveal that a significant portion of indoor objects, such as furniture and pets, exhibit higher FPRs in images of females than in those of males. Conversely, outdoor objects, such as vehicles, tend to have higher misclassification rates in images of males. These findings suggest that the model more closely associates females with indoor objects. The FPR scores for different objects on the MSCOCO dataset are included in the appendix.

\begin{table}[ht]
    \adjustbox{max width=0.48\textwidth}{
    \centering
    \small
    \begin{tabular}{lrrr}
        \toprule
        Dataset & \textsc{Bias\textsubscript{VL}} & \textsc{Bias\textsubscript{VL}}(M\textsubscript{1},M\textsubscript{2}) & \textsc{Bias\textsubscript{VL}}(F\textsubscript{1},F\textsubscript{2}) \\ 
        \midrule
        MSCOCO & 1.434 & 0.278 & 0.359\\ 
        PASCAL-S & 0.763 & 0.341 & 0.381 \\ 
        \bottomrule
    \end{tabular}
    }\caption{\textsc{Bias\textsubscript{VL}} for MSCOCO and PASCAL-SENTENCE (PASCAL-S) Datasets without any intervention. M and F stand for ``male'' and ``female'' respectively. \textsc{Bias\textsubscript{VL}} values obtained in two sets of images with the same gender are significantly lower than the \textsc{Bias\textsubscript{VL}}  obtained from datasets divided by gender.}
    \label{tab:indirect_bias}
\end{table}

\subsection{Bias Understanding with Causal Mediation Analysis}

We conduct the causal mediation analysis on different modules to study their effect on the model bias. We find that the image module influences the model bias more than the text module and the fusion module. In addition, we show that the bias in the image and text modules are aligned -- they are showing similar gender bias tendencies rather than conflicting ones. At this stage, we used four Quadro RTX 6000 GPUs. Since we primarily focus on inference tasks rather than training or fine-tuning, the computational resource requirements are not intensive.
\begin{figure*} 
    \centering
    \begin{subfigure}[t]{0.245\textwidth}
        \centering
        \includegraphics[width=\textwidth,height=2.7cm]{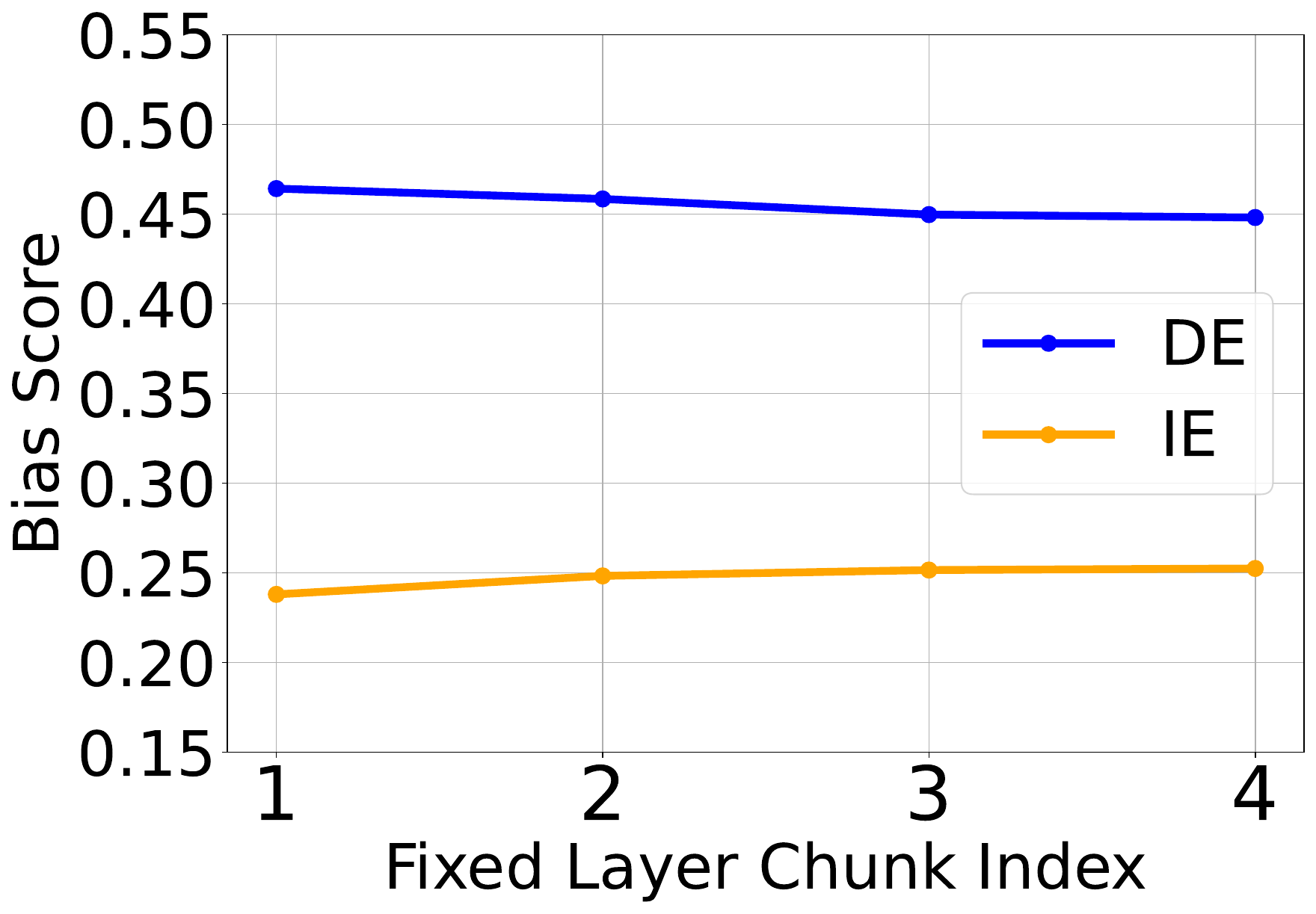}
        \caption{image encoder\\(COCO)}
        \label{fig:image_coco}
    \end{subfigure}
    \hfill
    \begin{subfigure}[t]{0.245\textwidth}
        \centering
        \includegraphics[width=\textwidth,height=2.7cm]{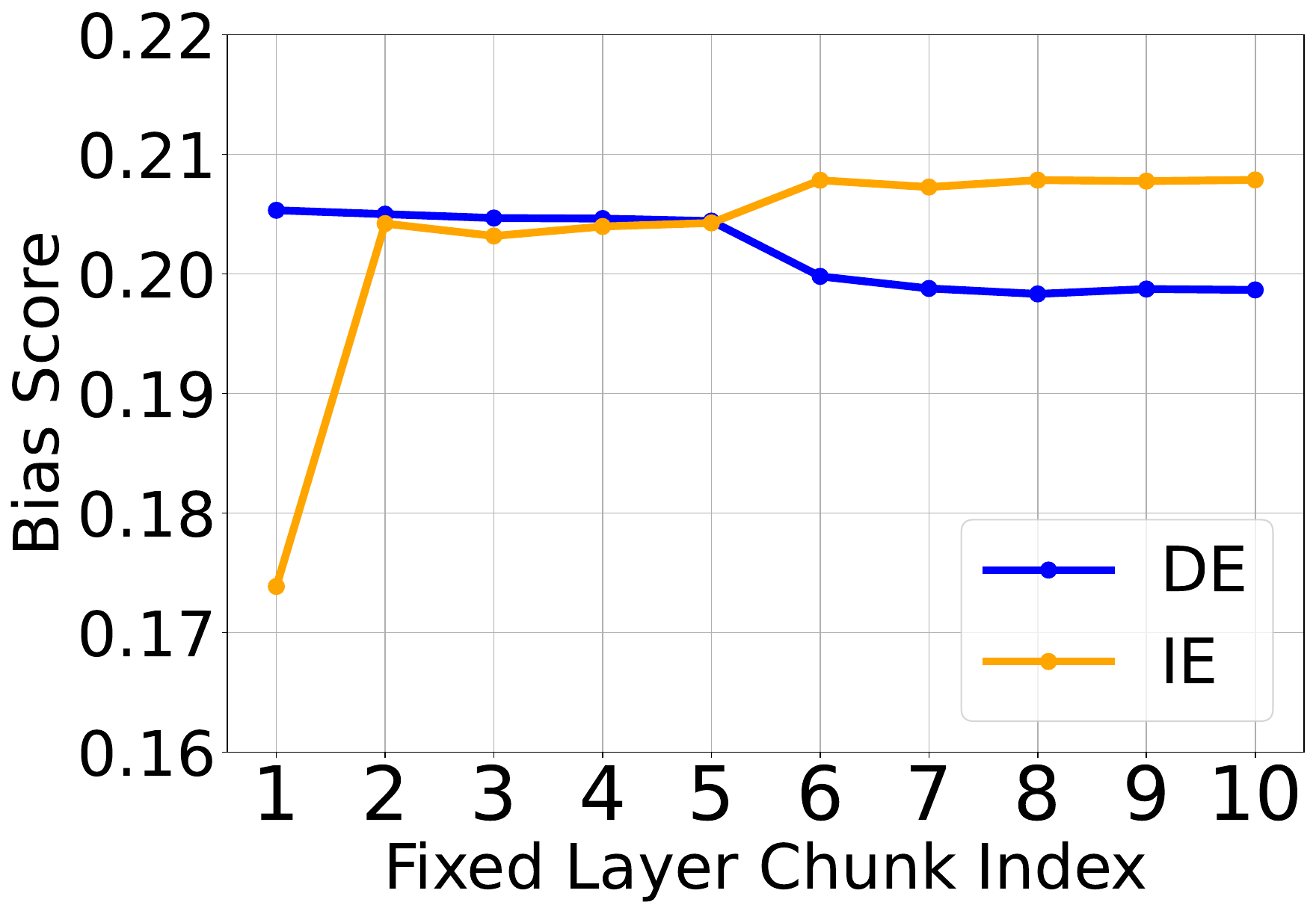}
        \caption{text encoder\\(COCO)}
        \label{fig:text_coco}
    \end{subfigure}
    \hfill
    \begin{subfigure}[t]{0.245\textwidth}
        \centering
        \includegraphics[width=\textwidth,height=2.7cm]{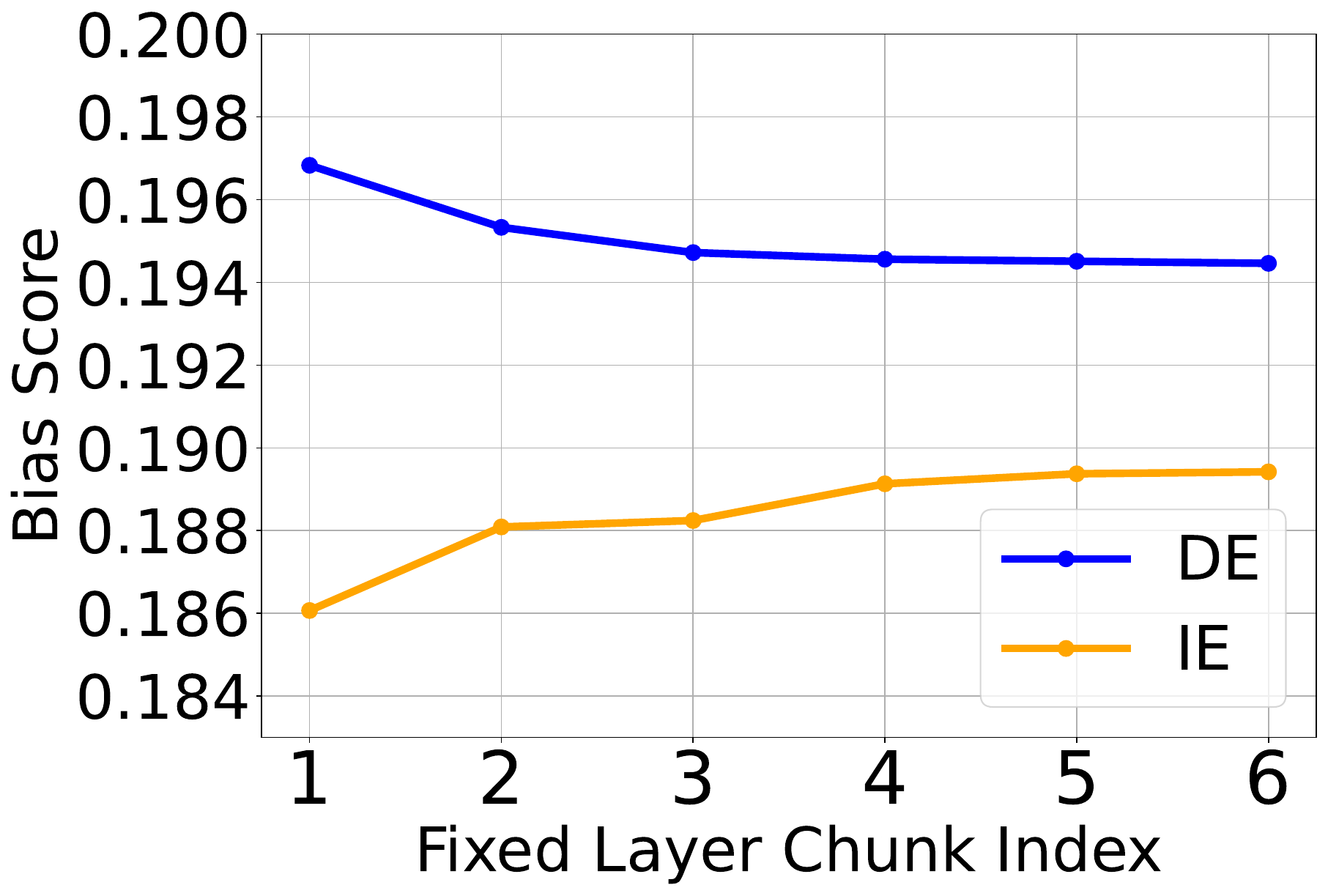}
        \caption{text part of deep-fusion \\encoder (COCO)}
        \label{fig:deepfusion_coco_text}
    \end{subfigure}
    \hfill
    \begin{subfigure}[t]{0.245\textwidth}
        \centering
        \includegraphics[width=\textwidth,height=2.65cm]{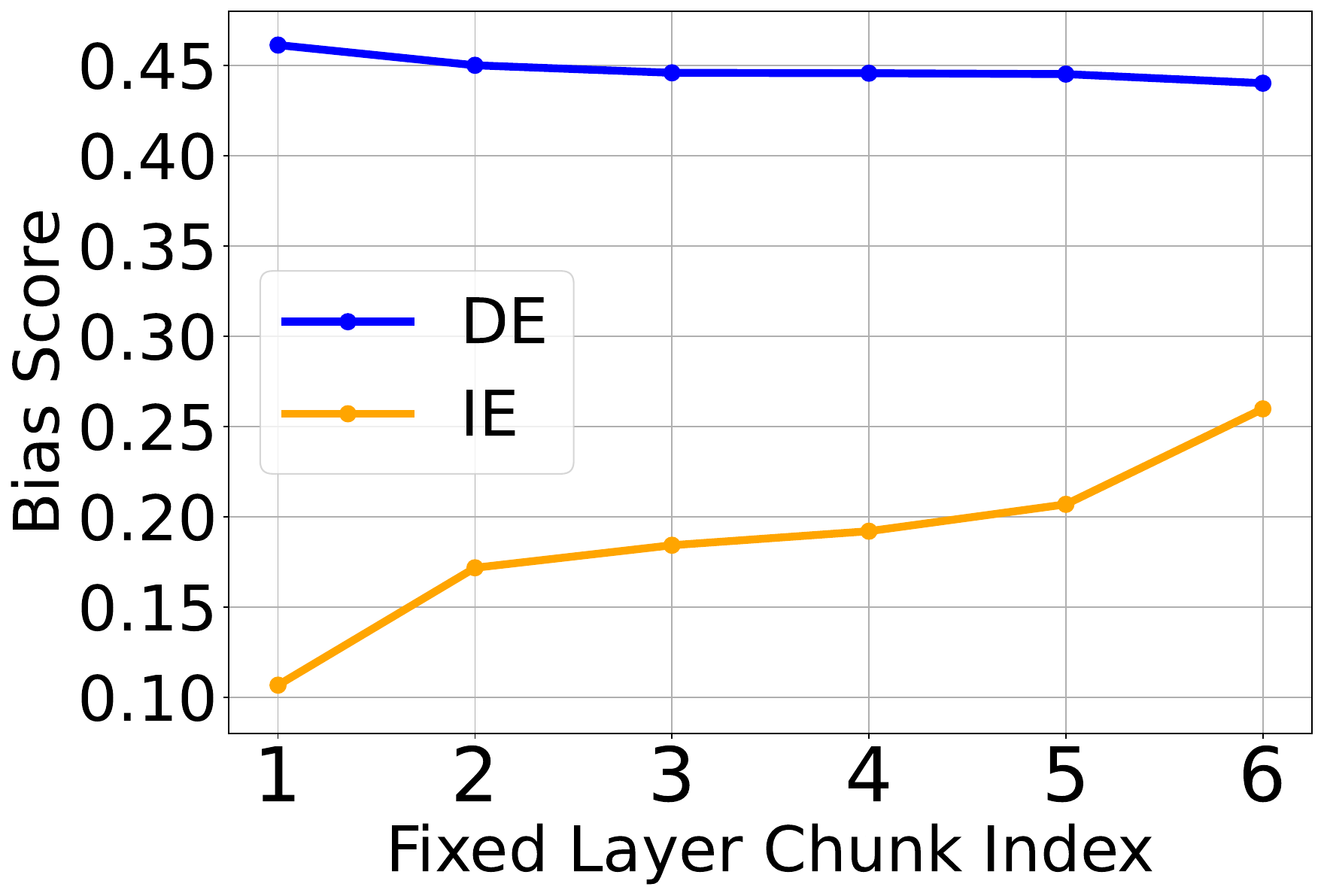}
        \caption{image part of deep-fusion encoder (COCO)}
        \label{fig:deepfusion_coco_image}
    \end{subfigure}
    \vspace{0.2cm} 
    \begin{subfigure}[t]{0.245\textwidth}
        \centering
        \includegraphics[width=\textwidth,height=2.7cm]{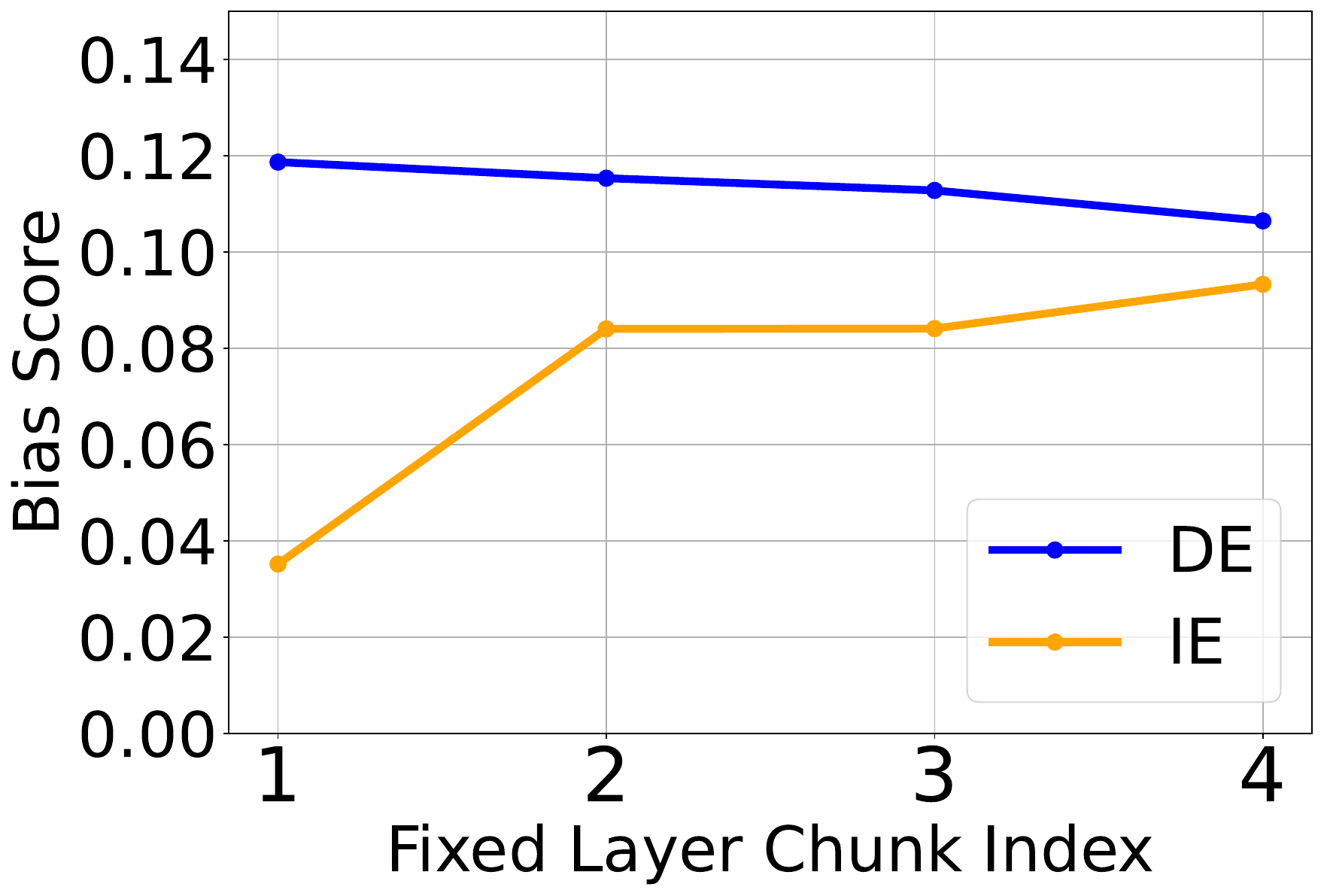}
        \caption{image encoder\\(PASCAL-S)}
        \label{fig:image_pascal}
    \end{subfigure}
    \hfill
    \begin{subfigure}[t]{0.245\textwidth}
        \centering
        \includegraphics[width=\textwidth,height=2.7cm]{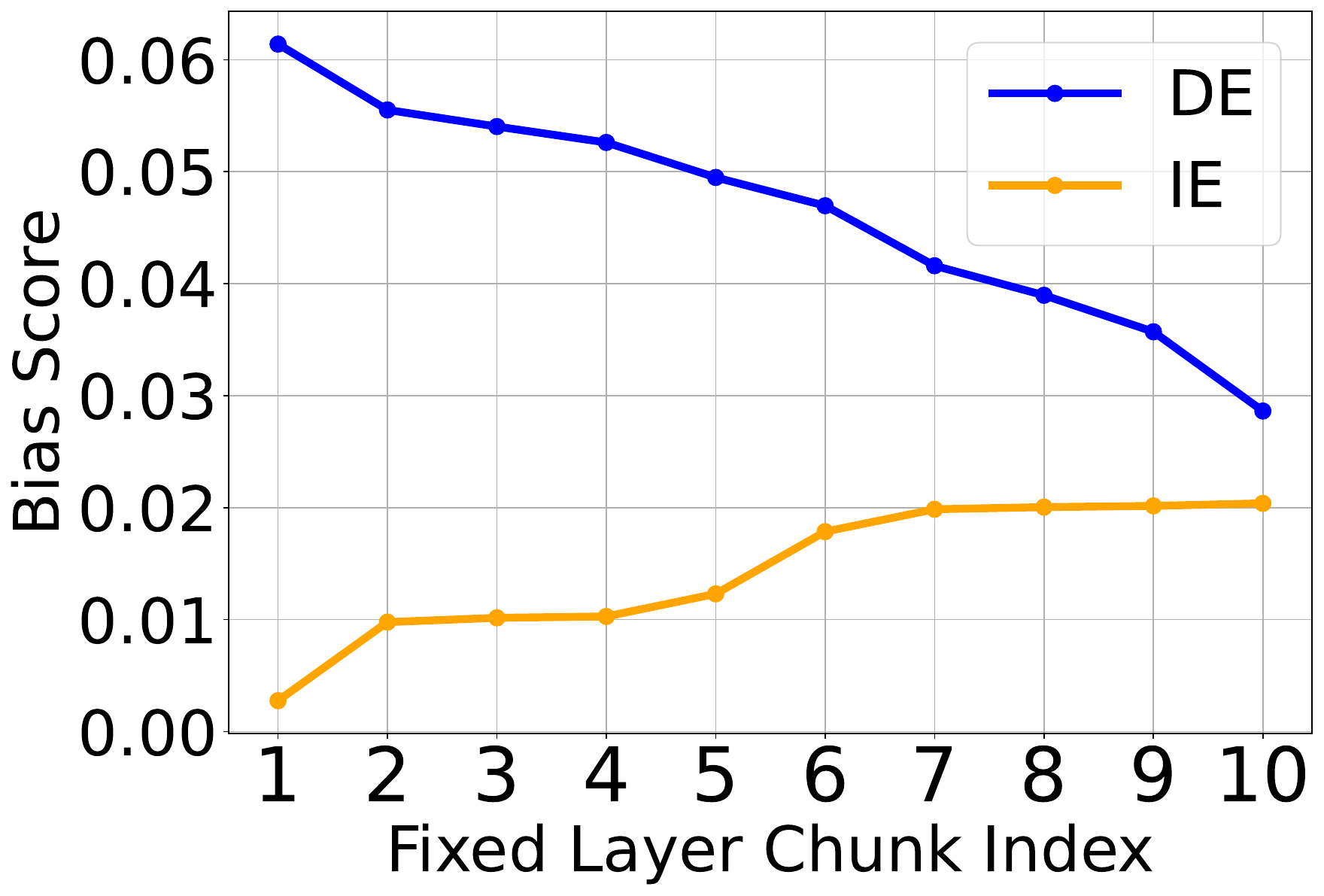}
        \caption{text encoder\\(PASCAL-S)}
        \label{fig:text_pascal}
    \end{subfigure}
    \hfill
    \begin{subfigure}[t]{0.245\textwidth}
        \centering
        \includegraphics[width=\textwidth,height=2.7cm]{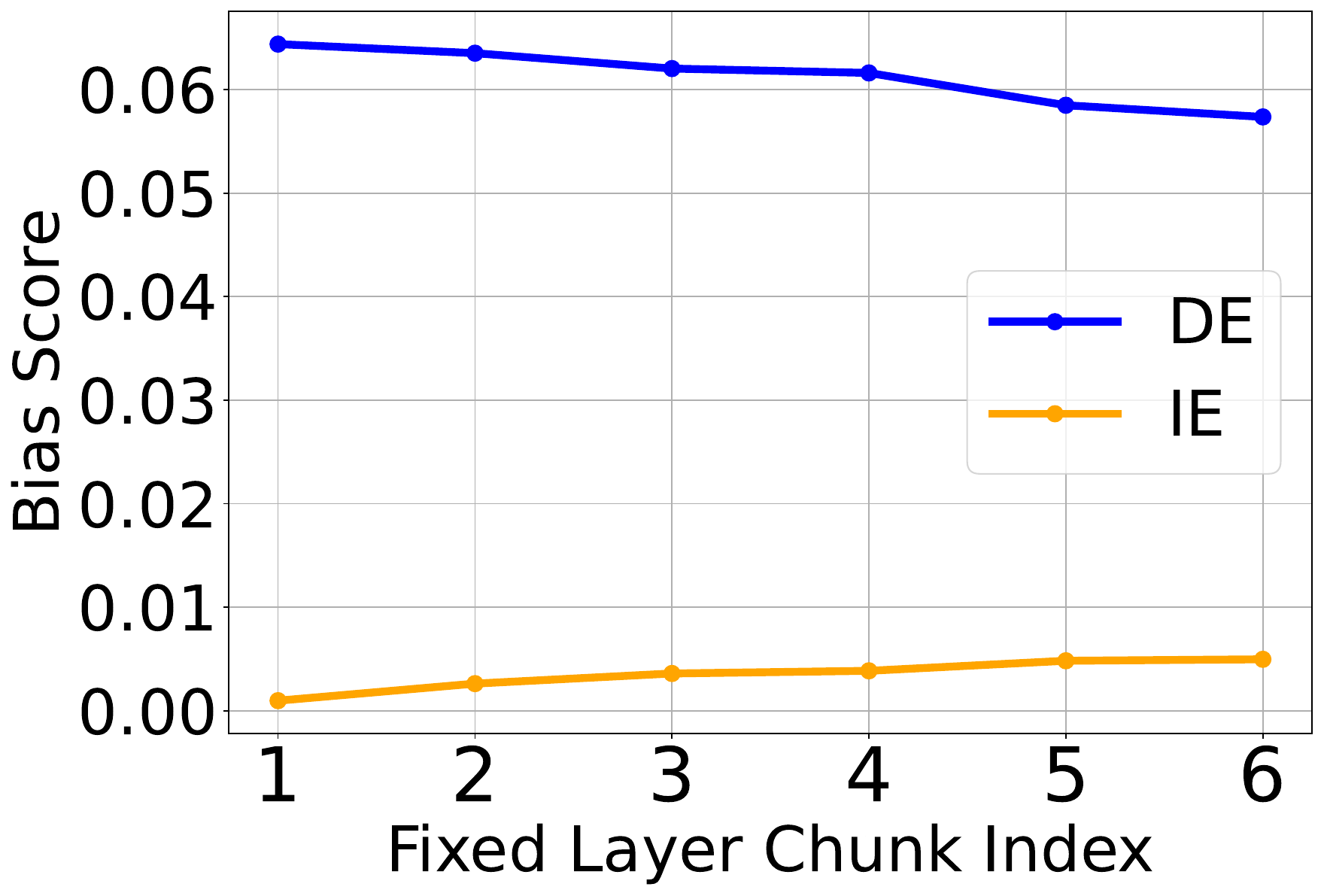}
        \caption{text part of deep-fusion \\encoder (PASCAL-S)}
        \label{fig:deepfusion_pascal_text}
    \end{subfigure}
    \hfill
    \begin{subfigure}[t]{0.245\textwidth}
        \centering
        \includegraphics[width=\textwidth,height=2.7cm]{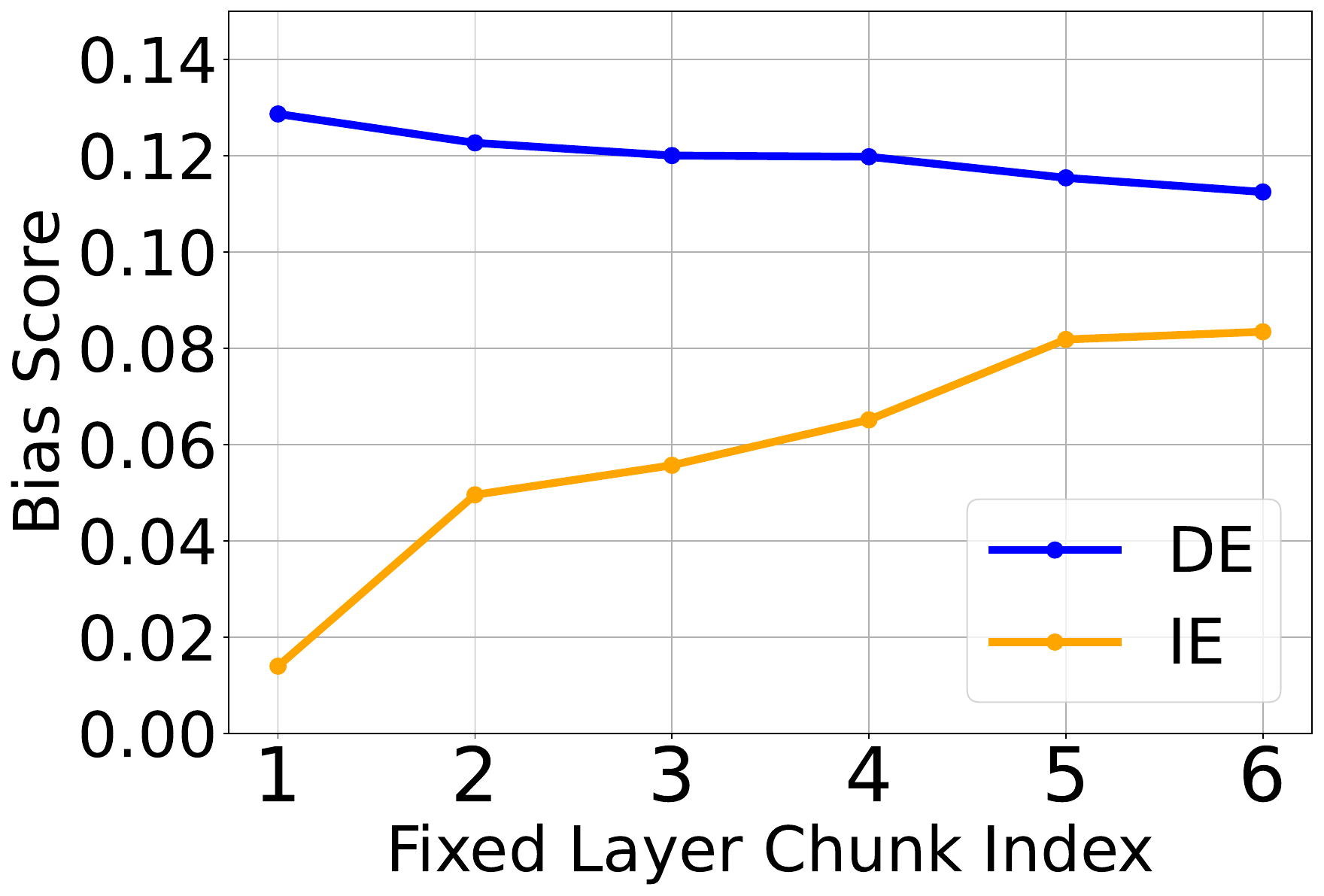}
        \caption{image part of deep-fusion encoder (PASCAL-S)}
        \label{fig:deepfusion_pascal_image}
    \end{subfigure}
    \caption{
    Causal mediation analysis of bias on the COCO and PASCAL-S (PASCAL-SENTENCE) datasets. Panels (a) and (e) show the DE (Direct Effect) and IE (Indirect Effect) for the image module; Panels (b) and (f) represent the DE and IE for the text module; Panels (c) and (g) illustrate the DE and IE for the text part of the deep-fusion encoder, and panels (d) and (h) for the image part of the deep-fusion encoder. The findings highlight that image features contribute more significantly to bias than text features, with the image module being the primary contributor to model bias.}
    
    \label{fig:causal_mediation}
\end{figure*}

\paragraph{Image encoder} Applying the \texttt{mask-gender} intervention to the input image module reduced the \textsc{Bias\textsubscript{VL}} to 0.967 for the MSCOCO dataset and to 0.664 for the PASCAL-SENTENCE dataset, representing reductions of approximately 32.57\% and 12.98\%, respectively. We employed the attention heads in the image encoder as the mediator to examine both the indirect effects of this model component and the direct effects of the \texttt{mask-gender} on predictions. Figure~\ref{fig:causal_mediation}\subref{fig:image_coco}  and Figure~\ref{fig:causal_mediation}\subref{fig:image_pascal} illustrate that employing more attention heads as mediators leads to an increase in the value of the indirect effect, while the direct effect diminishes. This supports an intuition that removing gender information from more layers in the image encoder weakens the model's dependency on latent correlations between gender in images and specific objects, thus mitigating gender bias in predictions. Furthermore, while interventions at the input level significantly impact final predictions, targeting the image encoder alone achieves about 53\% of the \texttt{mask-gender} effect.
\paragraph{Text encoder} Implementing a \texttt{replace-gender} intervention on the input text module reduced the \textsc{Bias\textsubscript{VL}} to 1.212 for the MSCOCO dataset and to 0.720 for the PASCAL-SENTENCE dataset, reductions of approximately 15.48\% and 5.64\%, respectively. We chose the attention heads within the text encoder as the mediator in this case. As shown in Figure~\ref{fig:causal_mediation}\subref{fig:text_coco}  and Figure~\ref{fig:causal_mediation}\subref{fig:text_pascal}, similar to the image encoder insights, removing gender information from multiple layers in the text encoder substantially decreases the model’s reliance on latent correlations between gender in text and specific objects, thereby reducing prediction biases. The \texttt{replace-gender} intervention led to a smaller reduction in bias compared to \texttt{mask-gender}, emphasizing the more substantial role of images in generating gender bias relative to text. This outcome is likely influenced by the simplistic structure of the input text used in our study, which adheres to the format described in original GLIP experiments~\cite{li2022grounded}, separating each category with a period, resulting in less complex text features than image features. Language models typically capture basic features such as syntactic structures at shallow layers and more complex semantic information at deeper layers, correlating with the significant changes in \textsc{Bias\textsubscript{VL}} observed at the sixth layer.

\paragraph{Deep fusion encoder} To further validate whether image features contribute more to bias generation than text features, we utilized the attention heads in the deep fusion encoder as the mediator, adjusting the attention heads' parameters in the states of either \texttt{mask-gender} intervention or \texttt{replace-gender} intervention. The results displayed in Figure~\ref{fig:causal_mediation}\subref{fig:deepfusion_coco_image} and Figure~\ref{fig:causal_mediation}\subref{fig:deepfusion_coco_text} show that for the MSCOCO dataset, the indirect effects from \texttt{mask-gender} and \texttt{replace-gender} through the deep fusion encoder are up to 0.260 and 0.189, respectively, reducing the \textsc{Bias\textsubscript{VL}} by approximately 18.13\% and 13.18\%. For the PASCAL-Sentence dataset, the reductions are 10.80\% and 0.53\%,  respectively (Figure~\ref{fig:causal_mediation}\subref{fig:deepfusion_pascal_image} and Figure~\ref{fig:causal_mediation}\subref{fig:deepfusion_pascal_text}). These findings reaffirm our conclusion that image features play a more substantial role in bias generation than text features. They also suggest that even though the deep fusion module does not extract features directly from images and text, the interactive updating process between text and image features significantly influences bias generation, accounting for approximately 55.7\% of the effect observed with the encoder alone.

\begin{figure}[h] 
    \centering
    \begin{subfigure}[b]{0.237\textwidth}
        \centering
        \includegraphics[width=\textwidth]{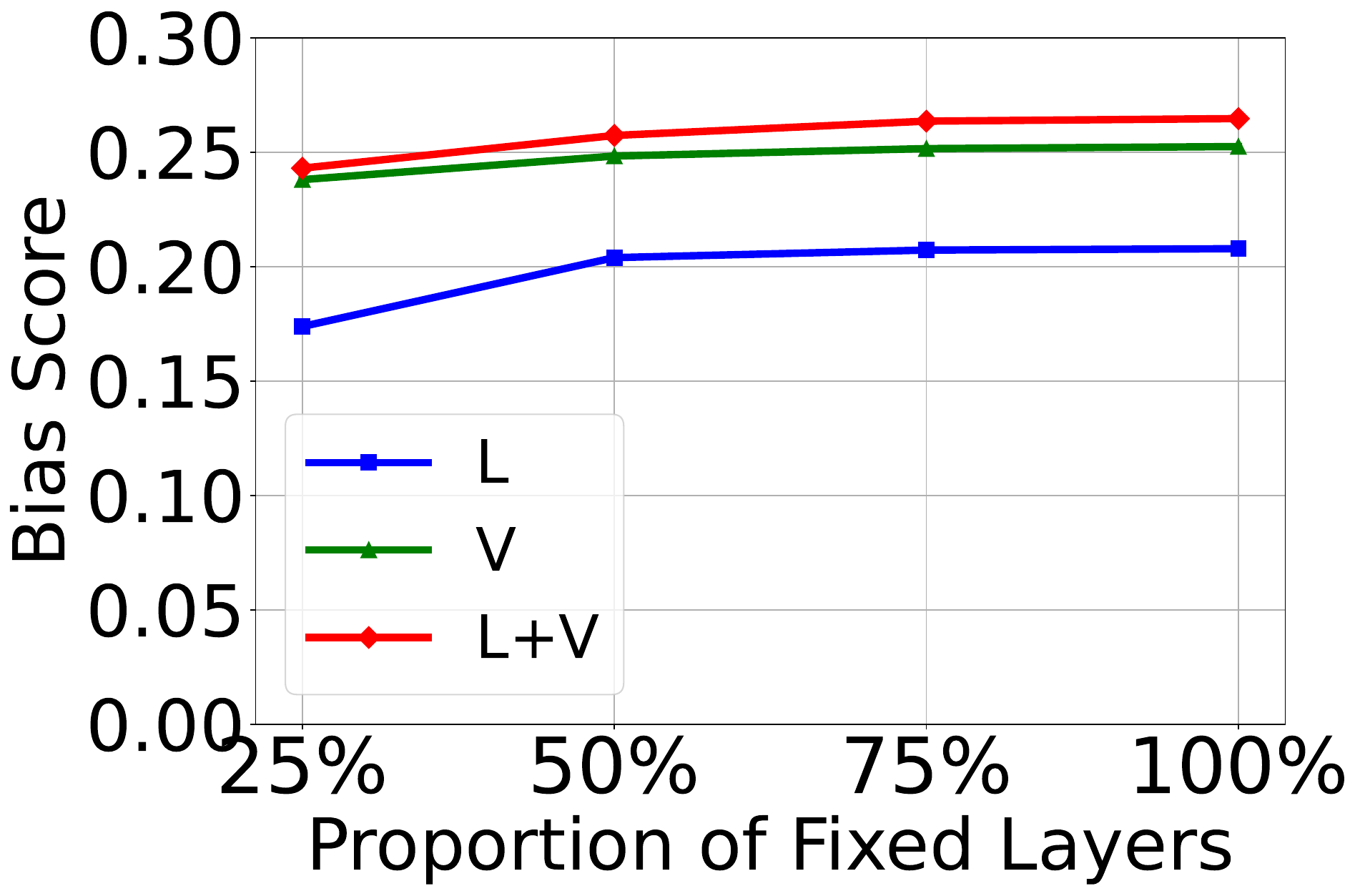}
        \caption{MSCOCO}
        \label{fig:coco_comparison}
    \end{subfigure}
    \hfill
    \begin{subfigure}[b]{0.237\textwidth}
        \centering
        \includegraphics[width=\textwidth]{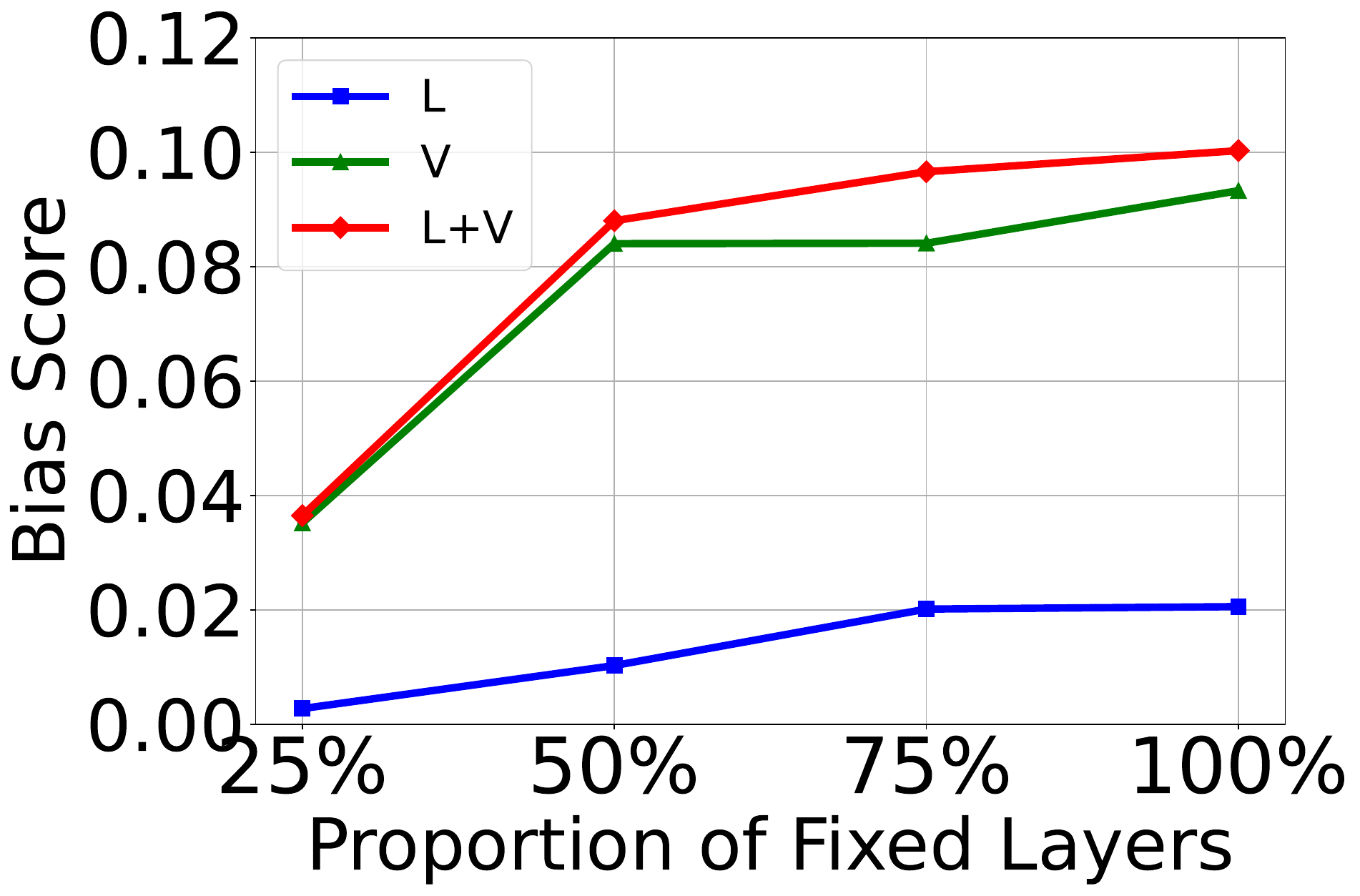}
        \caption{PASCAL-SENTENCE}
        \label{fig:pascal_comparison}
    \end{subfigure}
    \caption{Comparison of Bias Reduction Across Modalities with Interventions in Vision (V) and Language Modules (L) on the MSCOCO and PASCAL-SENTENCE datasets. \textbf{V} represents the results of interventions in the vision modality, \textbf{L} represents the results of interventions in the language modality, and \textbf{L+V} represents the results of simultaneous interventions in both the vision and language modalities. The contributions to bias from the two modalities are aligned and non-conflicting. Intervening simultaneously in both the visual and language modalities results in a greater reduction of bias compared to interventions in any single modality alone.}
    \label{fig:modality_comparison}
\end{figure}

\paragraph{Interventions comparison}
Multi-modal models consist of various interacting modules, each of which can learn distinct biases. However, the current literature does not thoroughly investigate whether these biases are aligned or disparate across different modules. In this section, we conduct an empirical analysis in VLMs to address this question. 
We simultaneously intervene in both the vision and language modalities. We apply \texttt{replace-gender} and \texttt{mask-gender} interventions to the input module and select a consistent proportion of attention heads in both the image encoder and text encoder as mediators. This setup allows us to observe changes in \textsc{Bias\textsubscript{VL}} and compare these with the changes induced by interventions in single modalities.
Figure~\ref{fig:modality_comparison}\subref{fig:coco_comparison} and Figure~\ref{fig:modality_comparison}\subref{fig:pascal_comparison} demonstrate that combined interventions on both images and text achieve greater bias reduction than interventions on either alone. However, the total reduction is not merely additive; the overall bias reduction is less than the sum of the individual contributions.

\begin{table*}[ht]
    \centering
    \begin{tabular}{lcc|cc|cc}
        \toprule
        & \multicolumn{2}{c|}{AP} & \multicolumn{2}{c|}{Bias} & \multicolumn{2}{c}{Bias Mitigated} \\
        \cmidrule(r){2-3} \cmidrule(r){4-5} \cmidrule(r){6-7}
        & MSCOCO & LVIS& MSCOCO & PASCAL-S & MSCOCO & PASCAL-S \\
        \midrule
        GLIP & 46.6 & 17.6 & 1.434 & 0.763 & 0 & 0 \\
        GLIP\_ImageFair & 46.2 & 17.3 & 1.118 & 0.694 & 22.036\% & 9.043\% \\
        GLIP\_TextFair & 46.6 & 17.6& 1.322 & 0.754 & 7.810\% & 1.180\% \\
        \bottomrule
    \end{tabular}
    \caption{Performance comparison of different methods. AP (Average Precision) is the metric used for zero-shot object detection. ``\texttt{GLIP}'' represents the original GLIP model, ``\texttt{GLIP\_ImageFair}'' denotes the model with bias mitigation implemented in the image encoder, and ``\texttt{GLIP\_TextFair}'' refers to the model with bias mitigation applied in the text encoder. Intervention in the image encoder is more effective than the text encoder in reducing the bias score without significant performance loss.}
    
    \label{tab:mitigation_results}
\end{table*}

\section{Bias Mitigation Method}
Based on our experimental results, image features contribute most significantly to gender bias and the image encoder has a more pronounced impact on bias compared to the text encoder and deep-fusion encoder. Therefore, our intuition is that focusing on reducing gender representation in the image encoder will effectively reduce bias, especially when facing a computation budget. We use the bias mitigation achieved from the text encoder as a baseline, then focus on reducing bias from the image encoder and compare the results with the baseline.

\paragraph{Text Encoder} 
For the text encoder, we aim to blur the gender representation in text features. We modify the structure of the text encoder to first identify gender-related terms \textit{(man, woman, men, women, male, female, males, females)} in the input text. A new sentence is generated by replacing these gendered terms with their corresponding anti-gender terms ( i.e., \textit{man} to \textit{woman}, \textit{male} to \textit{female}). The text encoder's output features are the average of the original sentence's text features and the anti-gender sentence's text features. Since the only difference between the two sentences is the gendered terms, this approach effectively blurs gender representation within the text encoder. We then let model to make predictions and observe the reduction in \textsc{Bias\textsubscript{VL}}.

\paragraph{Image Encoder}
Similarly, for the image encoder, we aim to blur gender representation in image features. To achieve this, we incorporate MTCNN~\cite{zhang2016joint} as a face detector and MobileNet~\cite{sandler2018mobilenetv2} as a gender classifier into the existing image encoder framework. Both networks are lightweight, allowing their integration without significantly increasing the computational load during inference. When an image is input into the image encoder, the MTCNN~\cite{zhang2016joint} network first identifies potential faces and outlines them with bounding boxes. MobileNet~\cite{sandler2018mobilenetv2} then classifies the gender of the faces within these boxes.

We have prepared a male face image and a female face image in advance. Depending on the gender predicted by MobileNet~\cite{sandler2018mobilenetv2}, we replace the face in the bounding box with the corresponding pre-prepared anti-gender face image. The final image features output by the image encoder are an average of the original image features and the features of the newly introduced anti-gender face. This method effectively blurs the original gender representation in the image. Then we let the model to make predictions and observe the reduction in \textsc{Bias\textsubscript{VL}}.

\section{Experimental Setup of Bias Mitigation}

\paragraph{Model} We utilized the GLIP model, pre-trained on the O365, GoldG, CC3M, and SBU datasets ~\cite{li2022grounded}. In our setup, we incorporated an MTCNN~\cite{zhang2016joint} pre-trained on the Wider Face and CelebA datasets as a face detector within the image encoder. Additionally, we integrated a MobileNet ~\cite{sandler2018mobilenetv2} pre-trained on ImageNet to serve as a gender classifier.

\paragraph{Dataset} We evaluated the effectiveness of bias mitigation on the MSCOCO and PASCAL-SENTENCE datasets. To assess the model's object detection performance, we compared it with the original GLIP ~\cite{li2022grounded} on the MSCOCO and LVIS datasets using the AP (Average Precision) metric for zero-shot object detection.

\section{Results}

As indicated in Table~\ref{tab:mitigation_results}, blurring gender representations in the image encoder demonstrated significant bias mitigation on both the MSCOCO and PASCAL-SENTENCE datasets. The experimental findings suggest that obscuring gender information in the image encoder is more effective at reducing model bias compared to similar interventions in the text encoder. Our results show that by blurring gender representations in the image features within the image encoder, we effectively reduced model bias by approximately 22.036\% and 9.043\% on the MSCOCO and PASCAL-SENTENCE datasets, respectively, with minimal impact on model performance.

\section{Conclusion}
Vision-language models (VLMs) trained on large-scale image-text pair corpora are at risk of learning social biases from their training data. In this paper, we introduced a standardized framework incorporating causal mediation analysis to measure and understand the pathways through which model bias is generated and propagated within VLMs. We discovered that image features contribute significantly more to model bias than text features, and the contributions from the image encoder substantially exceed those from the text encoder and deep fusion encoder. Furthermore, the contributions to bias from different modalities reinforce each other. Subsequently, by focusing on the components that contribute most to bias, we efficiently reduced model bias. 

Our work provides a framework for measuring, understanding, and mitigating model bias, which, although utilized here within the realm of object detection, can be extended to a wide range of VLM tasks. We used gender bias as a case study to showcase our methodology, as many influential studies in the natural language processing and fairness fields have begun with the study of gender bias ~
\cite{bolukbasi2016man,zhao2019gender,zhao2017men,zhang2022counterfactually}. Gender bias is one of the most representative types of bias, and understanding it can provide valuable insights for studying other forms of bias. The frameworks and evaluation metrics we developed for gender bias are generalizable and can be extended to other biases. For instance, if ground truth labels about age are available in the image data, researchers can modify the text descriptions to include age-related terms and apply the same methodology to study age-related bias. However, our framework is primarily applicable to white-box models, as it requires interventions at the internal components of the model. A promising direction for future work would involve expanding our framework to encompass additional modalities such as audio or video. This expansion could further enhance our understanding of multimodal interactions and their impact on bias, as well as deepen insights into how different sensory inputs contribute to, or mitigate, biases in AI systems.

\section{Limitations}
Our work provides a framework for measuring, understanding, and mitigating model bias in vision-language models (VLMs), with broad applicability across various VLM tasks. However, our approach primarily applies to white-box models, as it requires interventions within the model’s internal components. Consequently, this limitation implies that our methods might not be directly applicable to scenarios where model internals are inaccessible or when dealing with black-box systems.

\bibliography{custom}

\begin{thebibliography}{38}
\providecommand{\natexlab}[1]{#1}

\bibitem[{Alayrac et~al.(2022)Alayrac, Donahue, Luc, Miech, Barr, Hasson, Lenc, Mensch, Millican, Reynolds et~al.}]{alayrac2022flamingo}
Jean-Baptiste Alayrac, Jeff Donahue, Pauline Luc, Antoine Miech, Iain Barr, Yana Hasson, Karel Lenc, Arthur Mensch, Katherine Millican, Malcolm Reynolds, et~al. 2022.
\newblock Flamingo: a visual language model for few-shot learning.
\newblock \emph{Advances in neural information processing systems}, 35:23716--23736.

\bibitem[{Antol et~al.(2015)Antol, Agrawal, Lu, Mitchell, Batra, Zitnick, and Parikh}]{antol2015vqa}
Stanislaw Antol, Aishwarya Agrawal, Jiasen Lu, Margaret Mitchell, Dhruv Batra, C~Lawrence Zitnick, and Devi Parikh. 2015.
\newblock Vqa: Visual question answering.
\newblock In \emph{Proceedings of the IEEE international conference on computer vision}, pages 2425--2433.

\bibitem[{Barrett et~al.(2019)Barrett, Kementchedjhieva, Elazar, Elliott, and S{\o}gaard}]{barrett2019adversarial}
Maria Barrett, Yova Kementchedjhieva, Yanai Elazar, Desmond Elliott, and Anders S{\o}gaard. 2019.
\newblock Adversarial removal of demographic attributes revisited.
\newblock In \emph{Proceedings of the 2019 Conference on Empirical Methods in Natural Language Processing and the 9th International Joint Conference on Natural Language Processing (EMNLP-IJCNLP)}, pages 6330--6335.

\bibitem[{Bolukbasi et~al.(2016)Bolukbasi, Chang, Zou, Saligrama, and Kalai}]{bolukbasi2016man}
Tolga Bolukbasi, Kai-Wei Chang, James~Y Zou, Venkatesh Saligrama, and Adam~T Kalai. 2016.
\newblock Man is to computer programmer as woman is to homemaker? debiasing word embeddings.
\newblock \emph{Advances in neural information processing systems}, 29.

\bibitem[{Dehdashtian et~al.(2024)Dehdashtian, Wang, and Boddeti}]{dehdashtian2024fairerclip}
Sepehr Dehdashtian, Lan Wang, and Vishnu~Naresh Boddeti. 2024.
\newblock Fairerclip: Debiasing clip's zero-shot predictions using functions in rkhss.
\newblock In \emph{International Conference on Learning Representations (ICLR)}.

\bibitem[{Howard et~al.(2024)Howard, Bhiwandiwalla, Fraser, and Kiritchenko}]{howard2024uncovering}
Phillip Howard, Anahita Bhiwandiwalla, Kathleen~C Fraser, and Svetlana Kiritchenko. 2024.
\newblock Uncovering bias in large vision-language models with counterfactuals.
\newblock \emph{arXiv preprint arXiv:2404.00166}.

\bibitem[{Jia et~al.(2021)Jia, Yang, Xia, Chen, Parekh, Pham, Le, Sung, Li, and Duerig}]{jia2021scaling}
Chao Jia, Yinfei Yang, Ye~Xia, Yi-Ting Chen, Zarana Parekh, Hieu Pham, Quoc Le, Yun-Hsuan Sung, Zhen Li, and Tom Duerig. 2021.
\newblock Scaling up visual and vision-language representation learning with noisy text supervision.
\newblock In \emph{International conference on machine learning}, pages 4904--4916. PMLR.

\bibitem[{Kim et~al.(2021)Kim, Son, and Kim}]{kim2021vilt}
Wonjae Kim, Bokyung Son, and Ildoo Kim. 2021.
\newblock Vilt: Vision-and-language transformer without convolution or region supervision.
\newblock In \emph{International conference on machine learning}, pages 5583--5594. PMLR.

\bibitem[{Kuo et~al.(2023)Kuo, Cui, Gu, Piergiovanni, and Angelova}]{kuo2023openvocabulary}
Weicheng Kuo, Yin Cui, Xiuye Gu, AJ~Piergiovanni, and Anelia Angelova. 2023.
\newblock \href {https://openreview.net/forum?id=MIMwy4kh9lf} {Open-vocabulary object detection upon frozen vision and language models}.
\newblock In \emph{The Eleventh International Conference on Learning Representations}.

\bibitem[{Kurita et~al.(2019)Kurita, Vyas, Pareek, Black, and Tsvetkov}]{kurita2019measuring}
Keita Kurita, Nidhi Vyas, Ayush Pareek, Alan~W Black, and Yulia Tsvetkov. 2019.
\newblock Measuring bias in contextualized word representations.
\newblock In \emph{Proceedings of the First Workshop on Gender Bias in Natural Language Processing}, pages 166--172.

\bibitem[{Kuznetsova et~al.(2020)Kuznetsova, Rom, Alldrin, Uijlings, Krasin, Pont-Tuset, Kamali, Popov, Malloci, Kolesnikov et~al.}]{kuznetsova2020open}
Alina Kuznetsova, Hassan Rom, Neil Alldrin, Jasper Uijlings, Ivan Krasin, Jordi Pont-Tuset, Shahab Kamali, Stefan Popov, Matteo Malloci, Alexander Kolesnikov, et~al. 2020.
\newblock The open images dataset v4: Unified image classification, object detection, and visual relationship detection at scale.
\newblock \emph{International journal of computer vision}, 128(7):1956--1981.

\bibitem[{Lee et~al.(2023)Lee, Bang, Lovenia, Cahyawijaya, Dai, and Fung}]{lee2023survey}
Nayeon Lee, Yejin Bang, Holy Lovenia, Samuel Cahyawijaya, Wenliang Dai, and Pascale Fung. 2023.
\newblock Survey of social bias in vision-language models.
\newblock \emph{arXiv preprint arXiv:2309.14381}.

\bibitem[{Li et~al.(2023)Li, Li, Savarese, and Hoi}]{li2023blip}
Junnan Li, Dongxu Li, Silvio Savarese, and Steven Hoi. 2023.
\newblock Blip-2: Bootstrapping language-image pre-training with frozen image encoders and large language models.
\newblock In \emph{International conference on machine learning}, pages 19730--19742. PMLR.

\bibitem[{Li et~al.(2022)Li, Zhang, Zhang, Yang, Li, Zhong, Wang, Yuan, Zhang, Hwang et~al.}]{li2022grounded}
Liunian~Harold Li, Pengchuan Zhang, Haotian Zhang, Jianwei Yang, Chunyuan Li, Yiwu Zhong, Lijuan Wang, Lu~Yuan, Lei Zhang, Jenq-Neng Hwang, et~al. 2022.
\newblock Grounded language-image pre-training.
\newblock In \emph{Proceedings of the IEEE/CVF conference on computer vision and pattern recognition}, pages 10965--10975.

\bibitem[{Li et~al.(2018)Li, Tao, Joty, Cai, and Luo}]{li2018vqa}
Qing Li, Qingyi Tao, Shafiq Joty, Jianfei Cai, and Jiebo Luo. 2018.
\newblock Vqa-e: Explaining, elaborating, and enhancing your answers for visual questions.
\newblock In \emph{Proceedings of the European Conference on Computer Vision (ECCV)}, pages 552--567.

\bibitem[{Lin et~al.(2014)Lin, Maire, Belongie, Hays, Perona, Ramanan, Doll{\'a}r, and Zitnick}]{lin2014microsoft}
Tsung-Yi Lin, Michael Maire, Serge Belongie, James Hays, Pietro Perona, Deva Ramanan, Piotr Doll{\'a}r, and C~Lawrence Zitnick. 2014.
\newblock Microsoft coco: Common objects in context.
\newblock In \emph{Computer vision--ECCV 2014: 13th European conference, zurich, Switzerland, September 6-12, 2014, proceedings, part v 13}, pages 740--755. Springer.

\bibitem[{Lu et~al.(2019)Lu, Batra, Parikh, and Lee}]{lu2019vilbert}
Jiasen Lu, Dhruv Batra, Devi Parikh, and Stefan Lee. 2019.
\newblock Vilbert: Pretraining task-agnostic visiolinguistic representations for vision-and-language tasks.
\newblock \emph{Advances in neural information processing systems}, 32.

\bibitem[{Pearl(2022)}]{pearl2022direct}
Judea Pearl. 2022.
\newblock Direct and indirect effects.
\newblock In \emph{Probabilistic and causal inference: the works of Judea Pearl}, pages 373--392.

\bibitem[{Plummer et~al.(2015)Plummer, Wang, Cervantes, Caicedo, Hockenmaier, and Lazebnik}]{7410660}
Bryan~A. Plummer, Liwei Wang, Chris~M. Cervantes, Juan~C. Caicedo, Julia Hockenmaier, and Svetlana Lazebnik. 2015.
\newblock \href {https://doi.org/10.1109/ICCV.2015.303} {Flickr30k entities: Collecting region-to-phrase correspondences for richer image-to-sentence models}.
\newblock In \emph{2015 IEEE International Conference on Computer Vision (ICCV)}, pages 2641--2649.

\bibitem[{Radford et~al.(2021)Radford, Kim, Hallacy, Ramesh, Goh, Agarwal, Sastry, Askell, Mishkin, Clark et~al.}]{radford2021learning}
Alec Radford, Jong~Wook Kim, Chris Hallacy, Aditya Ramesh, Gabriel Goh, Sandhini Agarwal, Girish Sastry, Amanda Askell, Pamela Mishkin, Jack Clark, et~al. 2021.
\newblock Learning transferable visual models from natural language supervision.
\newblock In \emph{International conference on machine learning}, pages 8748--8763. PmLR.

\bibitem[{Rashtchian et~al.(2010)Rashtchian, Young, Hodosh, and Hockenmaier}]{rashtchian2010collecting}
Cyrus Rashtchian, Peter Young, Micah Hodosh, and Julia Hockenmaier. 2010.
\newblock Collecting image annotations using amazon’s mechanical turk.
\newblock In \emph{Proceedings of the NAACL HLT 2010 workshop on creating speech and language data with Amazon’s Mechanical Turk}, pages 139--147.

\bibitem[{Robins(2003)}]{robins2003semantics}
James~M Robins. 2003.
\newblock Semantics of causal dag models and the identification of direct and indirect effects.
\newblock \emph{Highly structured stochastic systems}, pages 70--82.

\bibitem[{Robins and Greenland(1992)}]{robins1992identifiability}
James~M Robins and Sander Greenland. 1992.
\newblock Identifiability and exchangeability for direct and indirect effects.
\newblock \emph{Epidemiology}, 3(2):143--155.

\bibitem[{Sandler et~al.(2018)Sandler, Howard, Zhu, Zhmoginov, and Chen}]{sandler2018mobilenetv2}
Mark Sandler, Andrew Howard, Menglong Zhu, Andrey Zhmoginov, and Liang-Chieh Chen. 2018.
\newblock Mobilenetv2: Inverted residuals and linear bottlenecks.
\newblock In \emph{Proceedings of the IEEE conference on computer vision and pattern recognition}, pages 4510--4520.

\bibitem[{Seth et~al.(2023)Seth, Hemani, and Agarwal}]{seth2023dear}
Ashish Seth, Mayur Hemani, and Chirag Agarwal. 2023.
\newblock Dear: Debiasing vision-language models with additive residuals.
\newblock In \emph{Proceedings of the IEEE/CVF Conference on Computer Vision and Pattern Recognition}, pages 6820--6829.

\bibitem[{Srinivasan and Bisk(2022)}]{srinivasan-bisk-2022-worst}
Tejas Srinivasan and Yonatan Bisk. 2022.
\newblock \href {https://doi.org/10.18653/v1/2022.gebnlp-1.10} {Worst of both worlds: Biases compound in pre-trained vision-and-language models}.
\newblock In \emph{Proceedings of the 4th Workshop on Gender Bias in Natural Language Processing (GeBNLP)}, pages 77--85, Seattle, Washington. Association for Computational Linguistics.

\bibitem[{Vaswani et~al.(2017)Vaswani, Shazeer, Parmar, Uszkoreit, Jones, Gomez, Kaiser, and Polosukhin}]{vaswani2017attention}
Ashish Vaswani, Noam Shazeer, Niki Parmar, Jakob Uszkoreit, Llion Jones, Aidan~N Gomez, {\L}ukasz Kaiser, and Illia Polosukhin. 2017.
\newblock Attention is all you need.
\newblock \emph{Advances in neural information processing systems}, 30.

\bibitem[{Vig and Belinkov(2019)}]{vig2019analyzing}
Jesse Vig and Yonatan Belinkov. 2019.
\newblock Analyzing the structure of attention in a transformer language model.
\newblock In \emph{Proceedings of the 2019 ACL Workshop BlackboxNLP: Analyzing and Interpreting Neural Networks for NLP}, pages 63--76.

\bibitem[{Vig et~al.(2020)Vig, Gehrmann, Belinkov, Qian, Nevo, Singer, and Shieber}]{vig2020investigating}
Jesse Vig, Sebastian Gehrmann, Yonatan Belinkov, Sharon Qian, Daniel Nevo, Yaron Singer, and Stuart Shieber. 2020.
\newblock Investigating gender bias in language models using causal mediation analysis.
\newblock \emph{Advances in neural information processing systems}, 33:12388--12401.

\bibitem[{Wang and Russakovsky(2021)}]{wang2021directional}
Angelina Wang and Olga Russakovsky. 2021.
\newblock Directional bias amplification.
\newblock In \emph{International Conference on Machine Learning}, pages 10882--10893. PMLR.

\bibitem[{Yu et~al.(2022)Yu, Wang, Vasudevan, Yeung, Seyedhosseini, and Wu}]{yu2022coca}
Jiahui Yu, Zirui Wang, Vijay Vasudevan, Legg Yeung, Mojtaba Seyedhosseini, and Yonghui Wu. 2022.
\newblock \href {https://openreview.net/forum?id=Ee277P3AYC} {Coca: Contrastive captioners are image-text foundation models}.
\newblock \emph{Transactions on Machine Learning Research}.

\bibitem[{Zhang et~al.(2022{\natexlab{a}})Zhang, Zhang, Hu, Chen, Li, Dai, Wang, Yuan, Hwang, and Gao}]{zhang2022glipv2}
Haotian Zhang, Pengchuan Zhang, Xiaowei Hu, Yen-Chun Chen, Liunian Li, Xiyang Dai, Lijuan Wang, Lu~Yuan, Jenq-Neng Hwang, and Jianfeng Gao. 2022{\natexlab{a}}.
\newblock Glipv2: Unifying localization and vision-language understanding.
\newblock \emph{Advances in Neural Information Processing Systems}, 35:36067--36080.

\bibitem[{Zhang et~al.(2016)Zhang, Zhang, Li, and Qiao}]{zhang2016joint}
Kaipeng Zhang, Zhanpeng Zhang, Zhifeng Li, and Yu~Qiao. 2016.
\newblock Joint face detection and alignment using multitask cascaded convolutional networks.
\newblock \emph{IEEE signal processing letters}, 23(10):1499--1503.

\bibitem[{Zhang et~al.(2022{\natexlab{b}})Zhang, Wang, and Sang}]{zhang2022counterfactually}
Yi~Zhang, Junyang Wang, and Jitao Sang. 2022{\natexlab{b}}.
\newblock Counterfactually measuring and eliminating social bias in vision-language pre-training models.
\newblock In \emph{Proceedings of the 30th ACM International Conference on Multimedia}, pages 4996--5004.

\bibitem[{Zhao et~al.(2023)Zhao, Andrews, and Xiang}]{zhao2023men}
Dora Zhao, Jerone Andrews, and Alice Xiang. 2023.
\newblock Men also do laundry: Multi-attribute bias amplification.
\newblock In \emph{International Conference on Machine Learning}, pages 42000--42017. PMLR.

\bibitem[{Zhao et~al.(2019)Zhao, Wang, Yatskar, Cotterell, Ordonez, and Chang}]{zhao2019gender}
Jieyu Zhao, Tianlu Wang, Mark Yatskar, Ryan Cotterell, Vicente Ordonez, and Kai-Wei Chang. 2019.
\newblock Gender bias in contextualized word embeddings.
\newblock In \emph{Proceedings of the 2019 Conference of the North American Chapter of the Association for Computational Linguistics: Human Language Technologies, Volume 1 (Long and Short Papers)}, pages 629--634.

\bibitem[{Zhao et~al.(2017)Zhao, Wang, Yatskar, Ordonez, and Chang}]{zhao2017men}
Jieyu Zhao, Tianlu Wang, Mark Yatskar, Vicente Ordonez, and Kai-Wei Chang. 2017.
\newblock Men also like shopping: Reducing gender bias amplification using corpus-level constraints.
\newblock In \emph{Proceedings of the 2017 Conference on Empirical Methods in Natural Language Processing}, pages 2979--2989.

\bibitem[{Zhou et~al.(2022)Zhou, Lai, and Jiang}]{zhou-etal-2022-vlstereoset}
Kankan Zhou, Eason Lai, and Jing Jiang. 2022.
\newblock {VLS}tereo{S}et: A study of stereotypical bias in pre-trained vision-language models.
\newblock In \emph{Proceedings of the 2nd Conference of the Asia-Pacific Chapter of the Association for Computational Linguistics and the 12th International Joint Conference on Natural Language Processing (Volume 1: Long Papers)}.

\end{thebibliography}

\appendix

\section{Appendix}
\label{sec:appendix}
\begin{figure*}[ht]
  \centering
  \includegraphics[width=0.8\textwidth]{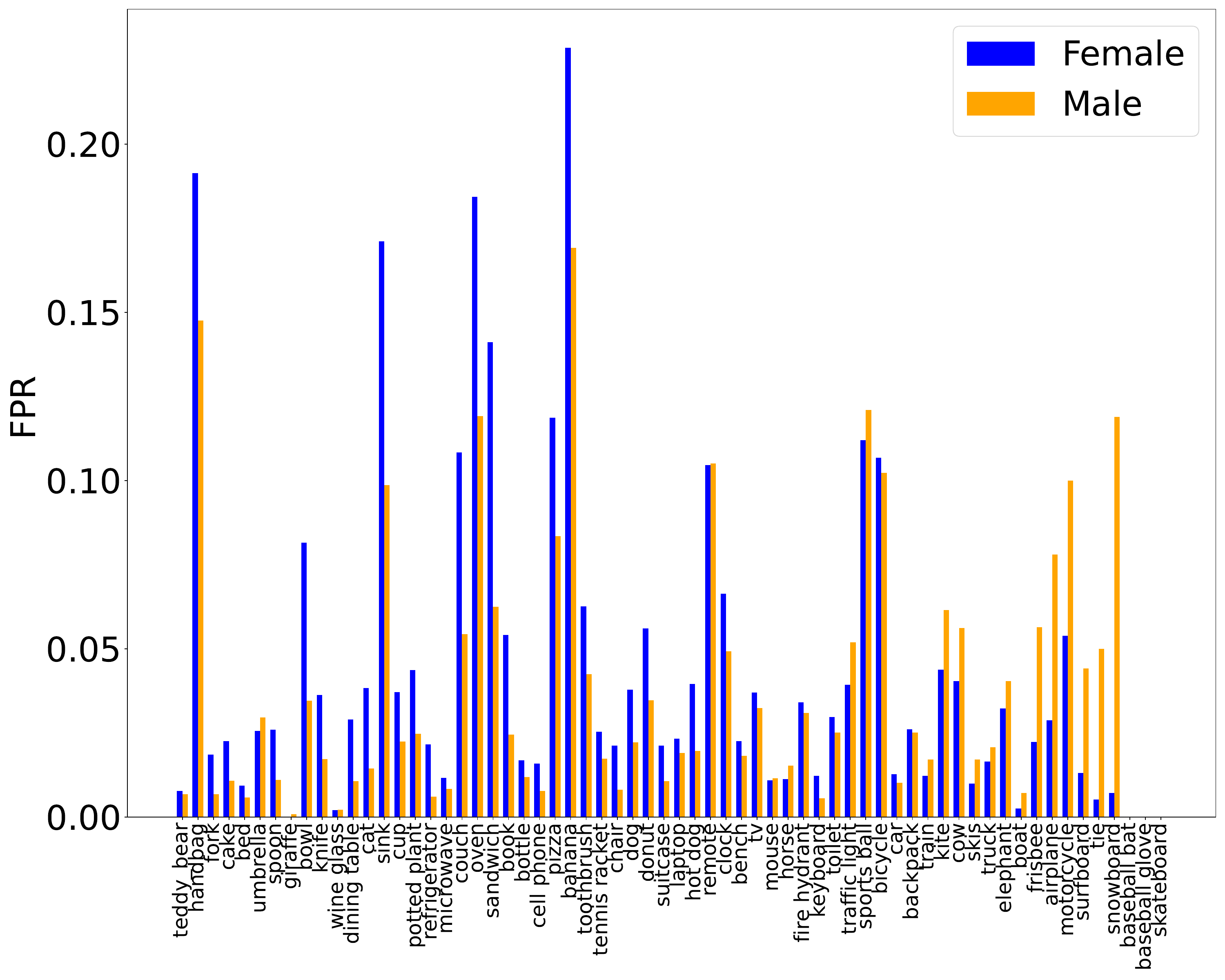}
  \caption{False Positive Rate (FPR) for various objects in the MSCOCO dataset. For most indoor objects, the FPR is higher in images of females than in those of males; conversely, for most outdoor objects such as vehicles, the FPR is higher in images of males than in those of females. These results indicate that females correlate more closely with indoor objects than males.}
  \label{fig:coco_fpr}
\end{figure*}

\end{document}